\documentclass[letterpaper]{article} 
\usepackage{aaai2026}  
\usepackage{times}  
\usepackage{helvet}  
\usepackage{courier}  
\usepackage[hyphens]{url}  
\usepackage{graphicx} 
\def\UrlFont{\rm}  
\usepackage{natbib}  
\usepackage{caption} 
\usepackage{algorithm}
\usepackage{algorithmic}
\usepackage{comment}

\usepackage{xcolor}
\usepackage{multirow}
\usepackage{booktabs}
\usepackage{verbatim}
\usepackage{subcaption}

\usepackage{amsmath}
\usepackage{amssymb} 
\usepackage{verbatim}

\usepackage{newfloat}
\usepackage{listings}
\DeclareCaptionStyle{ruled}{labelfont=normalfont,labelsep=colon,strut=off} 
\floatstyle{ruled}
\newfloat{listing}{tb}{lst}{}
\floatname{listing}{Listing}
\title{Debiased Dual-Invariant Defense for Adversarially Robust Person \\ Re-Identification}

\author{
    Yuhang Zhou\textsuperscript{\rm 1},
    Yanxiang Zhao\textsuperscript{\rm 1},
    Zhongyun Hua\textsuperscript{\rm 1}\thanks{Corresponding author},
    Zhipu Liu\textsuperscript{\rm 2}, \\
    Zhaoquan Gu\textsuperscript{\rm 1,3},
    Qing Liao\textsuperscript{\rm 1,3} ,
    Leo Yu Zhang\textsuperscript{\rm 4}
}

\affiliations{
    \textsuperscript{\rm 1} School of Computer Science and Technology, Harbin Institute of Technology, Shenzhen, China\\
    \textsuperscript{\rm 2} School of Computer Science and Engineering, Chongqing University of Technology, China\\
    \textsuperscript{\rm 3} Peng Cheng Laboratory, China\\
    \textsuperscript{\rm 4} School of Information and Communication Technology, Griffith University,  Australia\\

    23B951015@stu.hit.edu.cn, 24S151088@stu.hit.edu.cn,
    huazhongyun@hit.edu.cn, zpliu@cqut.edu.cn, guzhaoquan@hit.edu.cn, liaoqing@hit.edu.cn, leo.zhang@griffith.edu.au
}

\usepackage{bibentry}

\begin{document}

\maketitle

\begin{abstract}
Person re-identification (ReID) is a fundamental task in many real-world applications such as pedestrian trajectory tracking. However, advanced deep learning-based ReID models are highly susceptible to adversarial attacks, where imperceptible perturbations to pedestrian images can cause entirely incorrect predictions, posing significant security threats. Although numerous  adversarial defense strategies have been proposed for classification tasks, their extension to metric learning tasks such as person ReID remains relatively unexplored. Moreover, the several existing defenses for person ReID fail to address the inherent unique challenges of adversarially robust ReID. In this paper, we systematically identify the challenges of adversarial defense in person ReID into two key issues: model bias and composite generalization requirements. To address them, we propose a debiased dual-invariant defense framework composed of two main phases. In the data balancing phase, we mitigate model bias using a diffusion-model-based data resampling strategy that promotes fairness and diversity in training data. In the bi-adversarial self-meta defense phase, we introduce a novel metric adversarial training approach incorporating farthest negative extension softening to overcome the robustness degradation caused by the absence of classifier. Additionally, we introduce an adversarially-enhanced self-meta mechanism to achieve dual-generalization for both unseen identities and unseen attack types. Experiments demonstrate that our method significantly outperforms existing state-of-the-art defenses.
\end{abstract}

\begin{links}
    \link{Code}{https://github.com/zchuanqi/DDDefense-ReID}
\end{links}

\section{Introduction}
Person re-identification (ReID)~\citep{zhou2023stochastic,ye2021deep,wei2021fine} aims to retrieve a specific identity (ID) of interest from an image gallery. This capability enables practical applications such as tracking a suspect's movement trajectory from a single photo. The rapid development of deep learning provides high-performance ReID solutions. However, deep neural networks are vulnerable to adversarial attacks~\citep{goodfellow2014explaining,madry2018towards,croce2020reliable}, where the addition of human-imperceptible perturbations to input data can cause the model to produce completely incorrect predictions. This vulnerability also extends to person ReID systems, raising even more serious and realistic security threats. For example, malicious actors may manipulate inputs to evade detection or tracking by ReID model-based surveillance systems.

\begin{figure}[t]
    \centering
    \begin{subfigure}[b]{0.42\linewidth} 
        \centering
        \includegraphics[width=\textwidth]{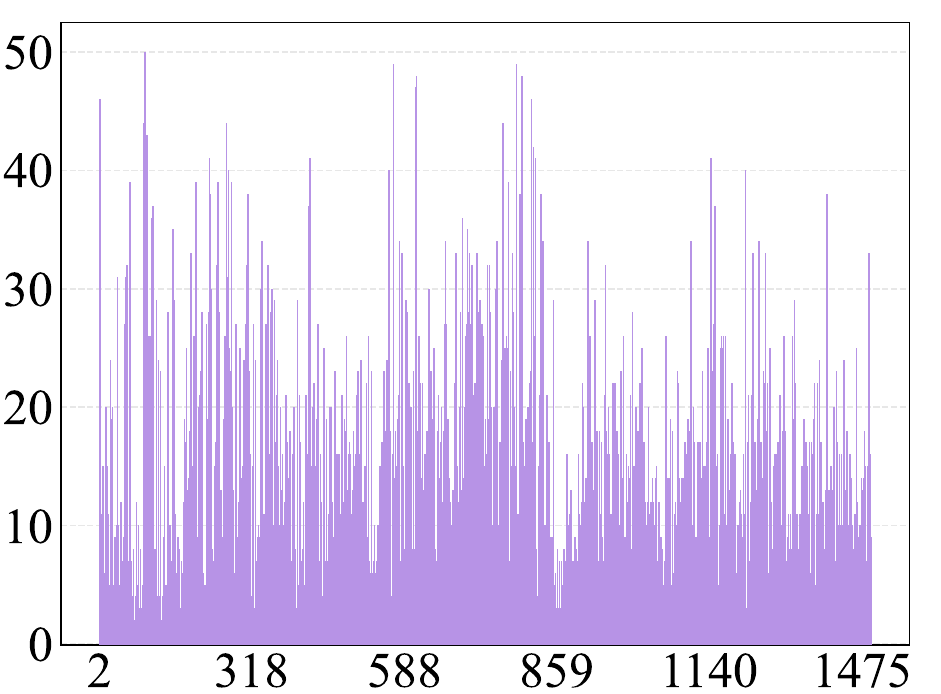} 
        \caption{Market-1501}
    \end{subfigure}
    \begin{subfigure}[b]{0.42\linewidth} 
        \centering
        \includegraphics[width=\textwidth]{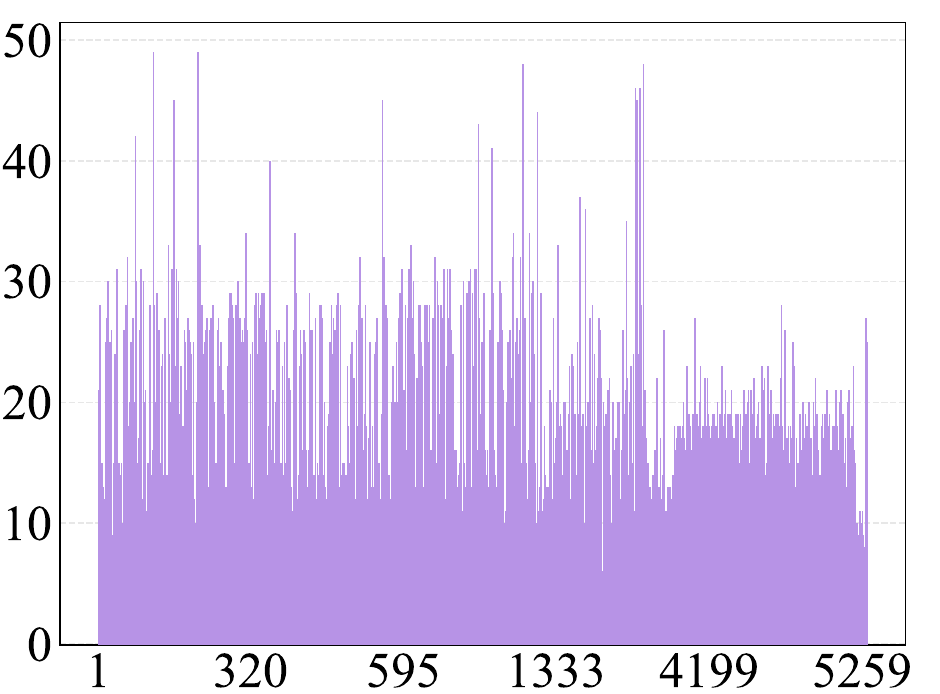} 
        \caption{DukeMTMC}
    \end{subfigure}
    \caption{Statistics of sample counts for each ID.}
\label{fig:data_number}
\end{figure}
\begin{figure}[t]
    \centering
    \begin{subfigure}[b]{0.42\linewidth} 
        \centering
        \includegraphics[width=\textwidth]{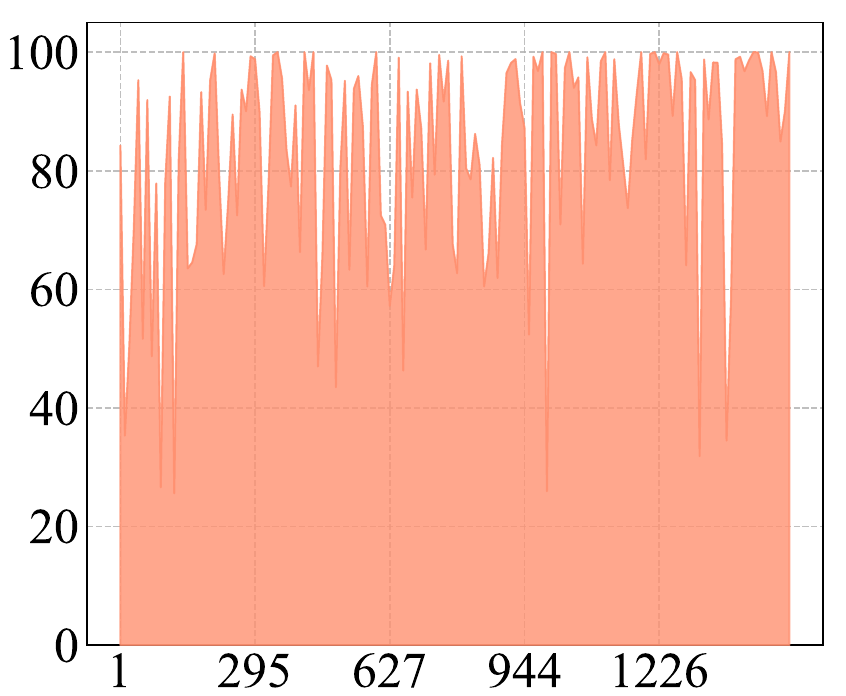} 
        \caption{The mAP of each ID for vanilla-trained model (avg.=84.13 and std.=18.78).}
    \end{subfigure}
    \hspace{0.2cm}
    \begin{subfigure}[b]{0.42\linewidth} 
        \centering
        \includegraphics[width=\textwidth]{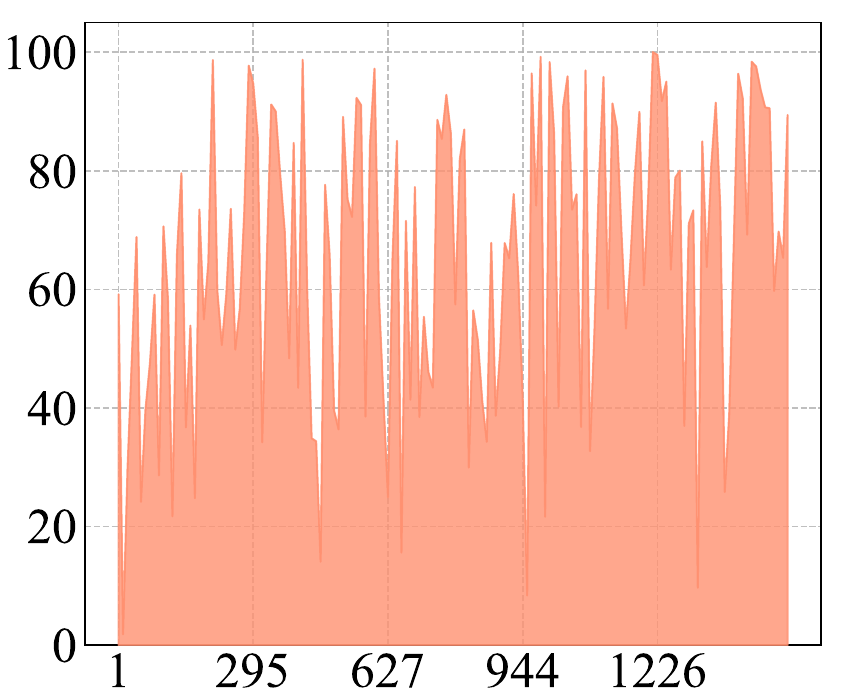} 
        \caption{The mAP of each ID for adversarially-trained model (avg.=66.10 and std.=23.40).}
    \end{subfigure}
    \caption{Biased model accuracy.}
\label{fig:biased_map}
\end{figure}
\begin{figure}[t]
  \centering
  \includegraphics[width=0.8\linewidth]{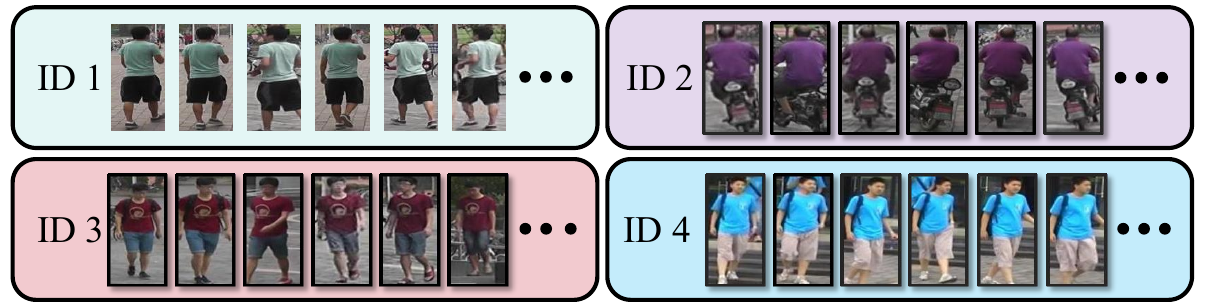}
  \caption{Partial adversarial examples from the original dataset illustrating the challenge of homogenization.}
  \label{fig:low_diversity}
\end{figure}
\begin{table}[t]
\fontsize{8}{8}\selectfont 
\centering
\caption{Validation of the hypothesis that ``partial robustness knowledge is distributed on the classifier".}
\setlength{\tabcolsep}{2mm}{
\begin{tabular}{cccc}
\toprule
Models                    & Defense                      & clean  & PGD    \\ \midrule
\multirow{2}{*}{ResNet18} & AT                           & 87.273 & 54.045 \\
                          & AT → fine-tuned(only classifier) & 88.113 & 50.752 \\ \cline{2-4}
\multirow{2}{*}{ResNet50} & AT                           & 88.786 & 53.076 \\
                          & AT → fine-tuned(only classifier) & 89.161 & 50.218  \\ \bottomrule
\end{tabular}}
\label{tab:veri_classifier}
\end{table}

Several adversarial attack methods targeting person ReID have been proposed~\cite{bai2020adversarial,wang2019advpattern,zheng2023u,bouniot2020vulnerability}. Early works focused on metric-based white-box attacks~\citep{bai2020adversarial,bouniot2020vulnerability}, which revealed the vulnerability of ReID models when full access to the model is available. Subsequently, black-box attack methods are developed, enabling adversarial manipulation without direct access to the target model~\citep{liu2023cat, zhang2020attacks, tsipras2018robustness, zhang2019theoretically, raghunathan2019adversarial}. In contrast to the research on adversarial attacks, defense strategies for person ReID remain relatively scarce. \citet{bai2020adversarial} propose an offline adversarial training method based on adversarial metric samples. \citet{bian2025learning} leverage virtual data to enhance the transferability of adversarial robustness. More recently, \citet{wei2024towards} develop a dynamic attack budget strategy  to enhance defense effectiveness. However,  these existing methods fail to consider the unique characteristics of person ReID, resulting in robustness degradation. In the supplementary material, we provide a detailed related works, including \textit{adversarial attacks, defenses}, and \textit{person ReID}.

In this work, we systematically re-examine the unique challenges of adversarial defense in person ReID and identify two key factors that contribute to the limitations of existing defense approaches:



\textbf{Model bias.} We attribute the model bias to two main issues. \textit{(i)} ReID datasets exhibit significant variation in inter-ID samples number, since the sample number for each ID depends on the frequency of its appearance under the camera. To demonstrate it, we report the per-identity sample distributions in two widely used datasets: DukeMTMC~\cite{zheng2015market} and Market-1501~\cite{zheng2017dukemtmc}. The results are shown in Figure~\ref{fig:data_number}. This imbalance leads to biased and unfair feature representations, as evidenced by the inter-ID accuracy variance of models trained using vanilla training (cross-entropy + triplet loss) and standard adversarial training~\cite{bouniot2020vulnerability}, shown in Figure~\ref{fig:biased_map}. \textit{(ii)} Datasets are typically extracted from video sequences, often resulting in high intra-ID redundancy and limited visual diversity, as illustrated in Figure~\ref{fig:low_diversity}. This homogeneity constrains the amount of useful information each identity provides during training, as visually similar frames contribute little to entropy or feature discrimination.


\textbf{Composite generalization requirements.} Compared to other common tasks (i.e., classification task), ReID task shows two unique generalization requirements. \textit{(i)} Adversarial training inevitably allocates partial robustness to the classifier which is unusable for ReID during testing, resulting in a decrease in robustness. To verify this issue, we adopt standard adversarial training to obtain a robust model A on CIFAR10. Then, we freeze the feature encoder and fine-tune the classifier on clean samples, resulting in robust model B with a vanilla classifier. The evaluation results of models A and B, as shown in Table~\ref{tab:veri_classifier}, indicate that the clean accuracy of the model after fine-tuning does not change largely, but robustness decreases significantly. This indicates that partial robustness knowledge is distributed on the classifier in adversarial training. Consequently, the feature encoder requires enhanced adversarial knowledge for generalized representation to reduce its dependency on the classifier's robustness. \textit{(ii)} Adversarial defense for person ReID is an open-set task while the attack strategies are also too diverse to enumerate during training. Consequently, adversarially robust ReID requires dual-dimensional adversarial generalization for both unseen IDs and unseen attacks, which poses more challenge than close-set task on which nowadays mainstream defense methods focus.

\begin{figure*}[t]
\centering
\includegraphics[width = 0.75\linewidth]{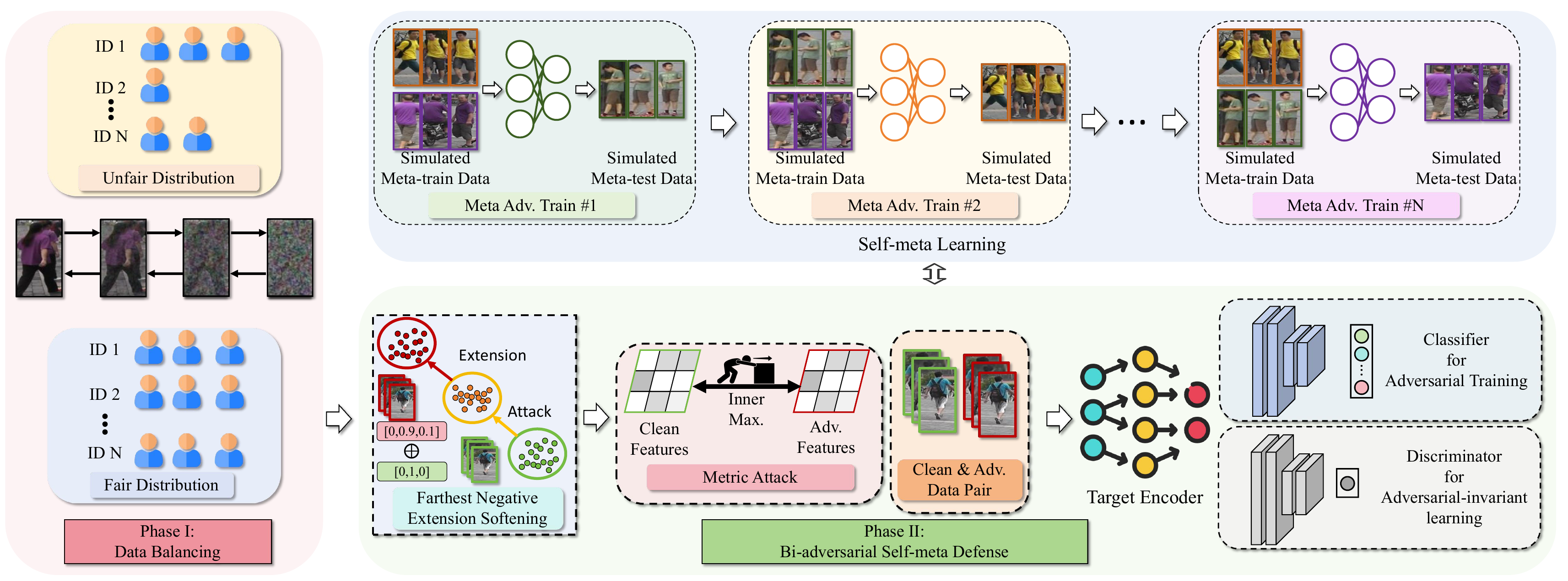}
\caption{Overview of our proposed method, which consists of data balancing and bi-adversarial self-meta defense.}
\label{fig:model}
\end{figure*}

To address the above challenges, we propose a novel composite adversarial defense framework comprising two key phases: a data balancing phase and a bi-adversarial self-meta defense phase. In the data balancing phase, we employ a diffusion-model-based augmentation strategy to optimize both inter-ID and intra-ID sample distributions. In the bi-adversarial self-meta defense phase, to address the robustness degradation caused by the absence of the classifier, we propose a novel metric adversarial training strategy based on farthest negative extension softening, which promotes robust representation learning by enhancing adversarial diversity. Additionally, we incorporate  label softening to mitigate the overfitting commonly caused by hard-label supervision. To further support generalization across both unseen identities and attack types, we introduce an adversarially-enhanced self-meta defense to extract dual-invariant features: inter-ID invariant features and adversarial-clean invariant features. Here, adversarial metric training and adversarial learning together form the bi-adversarial framework. Our main contributions can be summarized as follows:

\begin{itemize}
\item  We identify and empirically validate two fundamental challenges specific to adversarial defense in person ReID: model bias and the composite generalization requirement across both unseen identities and attack types.

\item  To address these challenges, we propose a novel composite defense framework for person ReID, integrating data balancing and bi-adversarial self-meta defense.

\item Experiments demonstrate that our method achieves state-of-the-art performance. Ablation studies confirm the contribution of each component, while transfer and interpretability evaluations highlight the framework's generalization capability and practical applicability.
\end{itemize}

\section{Methods}
\subsection{Preliminaries}
Let $E$, parameterized by $\theta_{E}$, denote the feature encoder, and $H$, parameterized by $\theta_{H}$, denote the classifier. Together, they form the complete model $G =H(E(\cdot))$, with parameters $\theta_{G} = \theta_{E} \cup \theta_{H}$. During training, both $E$ and $H$ are jointly optimized. However, during testing, only the feature encoder $E$ is used to extract features for retrieval. We define the target dataset distribution as $(x,y) \sim \mathcal{D}$, where $x$ represents the input samples and $y$ represents the  corresponding identity labels. Our method consists of two primary stages: data balancing and bi-adversarial self-meta defense.

\subsection{Data Balancing}
In this stage, we aim to address the fairness and diversity issues caused by data imbalance through a targeted data balancing strategy.

\textit{Data balancing via diffusion model.} To mitigate model bias caused by inter-ID data imbalance and limited intra-ID data diversity, we introduce a diffusion-model-based data balancing approach. Specifically, we train corresponding diffusion models on ReID datasets under an ID-conditional setting. For IDs with a small number of samples, we synthesize pseudo samples until a predefined sample threshold is met. The inter-ID data balancing process is formulated as follows:
{
\small
\begin{equation}
\begin{gathered}
\mathcal{D} \leftarrow \mathcal{D} \cup \left\{ {x}_{i}^{\text{pseudo},j} \mid i \in \mathcal{I},\ n_i < \delta_1,\ j = 1, \dots, \delta_1 - n_i \right\},
\end{gathered}
\end{equation}
}
where ${x}_{i}^{\text{pseudo}}$ denotes samples generated by the diffusion model with ID $i$, the set $\mathcal{I}$ denotes all IDs from the distribution $\mathcal{D}$, $n_i$ is the initial sample number of $i$ and $\delta_1$ is the threshold. Subsequently, IDs with a proportion of samples from any single camera exceeding a predefined threshold are deemed to lack diversity. For these IDs, we synthesize pseudo samples from other cameras. The intra-ID diversifying process is formulated as follows:
{
\small
\begin{equation}
\begin{gathered}
\mathcal{D}_{i'} \leftarrow \mathcal{D}_{i'} \cup \left\{ {x}_{{i'},c}^{\text{pseudo}} \mid c \in \mathcal{C} \backslash\{c_{i'}\}\right\},c_{i'}= \mathop{\arg\max}\limits_{c \in \mathcal{C}} \frac{n_{{i'},c}}{n_{{i'}}} \\
\end{gathered}
\end{equation}
}
where $\mathcal{D}_{i'}$ is the subset of $\mathcal{D}$ corresponding to ID $i'$, ${x}_{{i'},c}^{\text{pseudo}}$ denotes pseudo samples with ID ${i'}$ and camera $c$, the set $\mathcal{C}$ denotes all cameras of $\mathcal{D}$, $n_{{i'},c}$ is the sample number of ${i'}$ from $c$. ${i'}$ satisfies ${i'} \in \mathcal{I}$ and $\max\limits_{c \in \mathcal{C}}\frac{n_{{i'},c}}{n_{{i'}}}>\delta_2$, and $\delta_2$ is the predefined proportion threshold. The number of ${x}_{{i'},c}^{\text{pseudo}}$ is the initial mean value of each camera of $\mathcal{D}_{i'}$. In this implementation, we adopt the EDM framework~\cite{Karras2022edm}.

\subsection{Bi-adversarial Self-meta Defense}
In this stage, we propose a novel metric adversarial training based on a farthest negative extension softening method to address the robustness decreasing caused by the absence of the classifier with better attack diversity, and introduce an adversarially-enhanced self-meta defense to learning dual-invariant features: adversarial invariant features shared between clean samples and adversarial samples, and generalization invariant features shared between seen IDs and unseen IDs. Bi-adversarial framework consists of adversarial metric training and adversarial learning.

\textit{Adversarial metric training based on farthest negative extension softening.} The classifier $H$ is discarded during inference, where only features extracted by $E$ are used for computing distance metric. Consequently, current adversarial training methods for ReID commonly use metric variants of PGD~\cite{bouniot2020vulnerability}:
{
\small
\begin{equation}
\begin{split}
\hat{x}^{(0)} &= x + \eta ,\\
\hat{x}^{(n+1)} &= \Psi_{x}^{\epsilon} \Biggl( \hat{x}^{(n)} + \kappa \cdot \text{sign} \left( \frac{\partial \mathcal{L}_{\text{metric}}(\hat{x}^{(n)})}{\partial \hat{x}^{(n)}} \right) \Biggr) ,\\
\mathcal{L}_{\text{metric}}(x) &= \sum\limits_{x^{p} \in \mathcal{P}\cap\mathcal{A}} d\bigl( E(x), E(x^{p}) \bigr) \\
&\quad - \sum\limits_{x^{n} \in \mathcal{N}\cap\mathcal{A}} d\bigl( E(x), E(x^{n}) \bigr),
\end{split}
\end{equation}
}
where $x$ is the clean sample, $\hat{x}^{(n)}$ is the adversarial sample after the $n$-th iteration, $\eta$ is the initial perturbation, $\epsilon$ is the perturbation budget and $\kappa$ is the perturbation step size. $\textstyle\Psi_{x}^{\epsilon}$ is the clip function, which ensure that $\left\|\hat{x}^{(n+1)}-x\right\|_{\infty}\leq\epsilon$. The set $\mathcal{A}$ denotes other accessible samples, typically the current batch during training or the query set during inference. The set $\mathcal{P}$ denotes the samples with the same ID as $x$ and $\mathcal{N}$ denotes the samples belonging to the ID farthest from $x$. We use the euclidean distance as the distance metric $d(\cdot)$.

\begin{table*}[h]
\fontsize{7.5}{7}\selectfont  
\centering
\caption{White-box robustness results on ResNet50.}
\setlength{\tabcolsep}{1.5mm}{
\begin{tabular}{cclccccccc}
\toprule
\multirow{2}{*}{Model}     & \multirow{2}{*}{Dataset} & \multicolumn{1}{c}{\multirow{2}{*}{Defense}} & \multirow{2}{*}{Clean} & \multicolumn{2}{c}{FNA}  & \multicolumn{2}{c}{SMA}   & \multicolumn{2}{c}{IFGSM} \\ \cline{5-10} 
                           &                          & \multicolumn{1}{c}{}                         &                        & 8/255-16    & 10/255-16  & 8/255-16    & 10/255-16   & 8/255-16    & 10/255-16   \\ \midrule
\multirow{16}{*}{ResNet50} & \multirow{8}{*}{Market}  & Origin                                       & 78.49/92.01            & 0.20/0.17   & 0.18/0.14  & 0.27/0.26   & 0.20/0.11   & 1.25/1.95   & 1.09/1.66   \\
                           &                          & BitSqueezing                                 & 77.46/91.54            & 0.22/0.18   & 0.20/0.12  & 0.31/0.24   & 0.25/0.12   & 1.03/1.40   & 0.95/1.16   \\
                           &                          & MedianSmoothning2D                             & \textbf{78.55/91.95}            & 0.68/0.89   & 0.57/0.80  & 3.57/4.84   & 3.29/3.86   & 3.77/6.41   & 3.33/5.70   \\
                           &                          & BS+MS                                        & 77.22/90.06            & 1.33/2.08   & 1.08/1.84  & 6.98/10.15  & 5.90/8.67   & 6.31/11.58  & 5.57/10.27  \\
                           &                          & AMD                                          & 76.85/90.97            & 0.30/0.36   & 0.26/0.27  & 0.45/0.39   & 0.42/0.33   & 1.41/2.35   & 1.17/2.02   \\
                           &                          & Adv\_train                                  & 69.69/88.24            & 8.57/18.14  & 4.37/9.41  & 22.85/35.69 & 15.21/23.37 & 17.97/34.65 & 11.74/23.34 \\
                           &                          & DAS                                          & 69.79/88.39            & 12.70/24.85 & 7.25/14.52 & 32.14/49.05 & 24.37/38.69 & 22.33/39.79 & 15.90/30.93 \\
                           &                          & \textbf{Ours}                             & 68.50/88.21            & \textbf{31.99/55.17}   & \textbf{24.80/45.93 }   & \textbf{50.13/72.60}         & \textbf{45.96/67.34}         & \textbf{37.61/62.02}         & \textbf{31.20/54.22}         \\  \cline{3-10}
                           & \multirow{8}{*}{Duke}    & Origin                                       & 68.83/83.80            & 0.10/0.00   & 0.09/0.00  & 0.30/0.18   & 0.24/0.13   & 0.89/1.17   & 0.79/1.03   \\
                           &                          & BitSqueezing                                 & 68.36/83.03            & 0.13/0.18   & 0.12/0.13  & 0.64/0.58   & 0.50/0.27   & 1.44/2.15   & 1.25/1.66   \\
                           &                          & MedianSmoothning                             & 69.46/84.02            & 0.34/0.31   & 0.27/0.22  & 3.80/4.85   & 3.48/4.44   & 3.72/6.60   & 3.09/5.12   \\
                           &                          & BS+MS                                        & 68.04/83.21            & 0.58/0.63   & 0.44/0.36  & 6.36/8.62   & 5.07/6.87   & 5.45/8.98   & 4.58/7.50   \\
                           &                          & AMD                                          & \textbf{69.56/84.02}            & 0.10/0.00   & 0.10/0.00  & 0.46/0.36   & 0.38/0.31   & 1.08/1.53   & 0.97/1.39   \\
                           &                          & Adv\_train                                   & 57.29/75.31            & 7.47/13.64  & 3.86/7.68  & 18.46/29.44 & 13.28/20.65 & 16.64/29.94 & 12.07/22.31 \\
                           &                          & DAS                                          & 59.04/77.33            & 11.15/19.84 & 5.93/11.27 & 26.09/39.45 & 19.36/29.89 & 19.38/33.39 & 13.70/24.55 \\
                           &                          & \textbf{Ours}                                         & 55.29/75.81             & \textbf{26.02/43.49 }        & \textbf{20.76/36.54}        & \textbf{40.74/59.69}         & \textbf{37.50/55.83}         & \textbf{31.23/50.67}         & \textbf{26.89/45.29}         \\ \bottomrule
\end{tabular}}
\label{tab:white_box_resnet50}
\end{table*}
\begin{table*}[h]
\fontsize{7.5}{7}\selectfont  
\centering
\caption{Ablation analysis of our proposed model on ResNet50.}
\setlength{\tabcolsep}{1.5mm}{
\begin{tabular}{cccccccccc}
\toprule
\multirow{2}{*}{Models}    & \multirow{2}{*}{datasets} & \multirow{2}{*}{Modules} & \multirow{2}{*}{Clean} & \multicolumn{3}{c}{8/255-16}            & \multicolumn{3}{c}{10/255-16}           \\ \cline{5-10}
                           &                           &                          &                        & FNA         & SMA         & IFGSM       & FNA         & SMA         & IFGSM       \\ \midrule
\multirow{12}{*}{ResNet50} & \multirow{6}{*}{Market}   & Metric AT                       & 67.20/88.00            & 28.38/52.26 & 45.38/68.74 & 33.97/58.52 & 22.02/42.96 & 40.56/63.18 & 28.57/52.23 \\
                           &                           & +Diffusion model                    & 66.96/86.91            & 29.85/53.36 & 45.82/68.41 & 35.32/59.68 & 22.82/43.38 & 41.14/63.26 & 29.03/52.46 \\
                           &                           & +Adversarial learning                    & 67.81/88.21            & 30.34/53.77 & 48.10/70.72 & 35.70/60.27 & 23.21/44.21 & 43.58/65.17 & 29.31/52.29 \\
                           &                           & +Self-meta learning                   & 68.24/88.03            & 29.48/52.88 & 46.91/69.30 & 35.09/59.86 & 22.19/42.96 & 42.06/64.52 & 28.61/52.31 \\
                           &                           & +FNES                   & 68.29/88.07            & 30.98/54.45 & 49.25/71.11 & 37.16/61.49 & 24.64/44.83 & 44.07/65.97 & 30.54/53.15 \\
                           &                           & \textbf{All modules}                      & \textbf{68.50/88.21}   & \textbf{31.99/55.17} & \textbf{50.13/72.60} & \textbf{37.61/62.02} & \textbf{24.80/45.93} & \textbf{45.96/67.34} & \textbf{31.20/54.22} \\ \cline{2-10}
                           & \multirow{6}{*}{Duke}     & Metric AT                       & 54.47/74.55            & 23.76/39.27 & 40.40/57.99 & 27.04/44.21 & 17.41/30.97 & 35.98/52.42 & 21.55/36.58 \\
                           &                           & +Diffusion model                    & 55.15/74.87            & 24.82/42.48 & 39.44/57.14 & 30.10/49.31 & 18.67/35.45 & 35.50/53.46 & 24.93/43.67 \\
                           &                           & +Adversarial learning                     & 54.55/75.13            & 25.38/42.03 & 38.67/57.71 & 30.75/49.55 & 19.72/34.86 & 36.08/54.67 & 25.97/42.91 \\
                           &                           & +Self-meta learning                    & 54.88/74.96            & 25.12/43.02 & 38.65/58.57 & 30.11/49.69 & 19.53/35.02 & 35.87/54.30 & 25.16/43.60 \\
                           &                           & +FNES                   & 54.81/75.40            & 25.20/42.75 & 40.18/58.14 & 30.23/49.53 & 19.82/35.21 & 36.89/54.11 & 25.60/43.57 \\
                           &                           & \textbf{All modules}                     & \textbf{55.29/75.81}    & \textbf{26.02/43.49 }& \textbf{40.74/59.69} & \textbf{31.23/50.67} & \textbf{20.76/36.54} & \textbf{37.50/55.83} & \textbf{26.89/45.29} \\ \bottomrule
\end{tabular}}
\label{tab:abla}
\end{table*}

To address the robustness decreasing caused by the absence of the classifier, we introduce farthest negative extension softening (FNES) for robust representation learning with better attack diversity. Specifically, after attacking, we apply linear scaling to the final adversarial perturbation and soften the label of adversarial sample on the farthest negative class as:
{
\small
\begin{equation}
\begin{split}
x^{\text{temp}}&=x+\gamma\cdot\left ( \hat{x}-x \right ),\\
x^{\text{adv}}&=\omega x+(1-\omega)x^{\text{temp}},\\
y^{\text{adv}}&=\omega\phi(y,\lambda_1)+(1-\omega)\tau\bigl(\phi(y,\lambda_2),\upsilon \bigr),
\end{split}
\label{equation:mixup}
\end{equation}
}
where $x^{\text{temp}}$ is the temporary adversarial sample after linear scaling, $\hat{x}$ is the adversarial sample before FNES, $\gamma\geq1$ is a scaling factor, $\omega$ is a real value sampled from a uniform distribution $\mathcal{U}(a,b)$ with $a,b\in(0,1)$ and $x^{\text{adv}}$ is the final adversarial sample obtained by linearly mixing $x$ and $x^{\text{temp}}$. $\phi$ is the label-smoothing function~\cite{Szegedy2016labelsmoothing}, assigning $\lambda\in(0,1)$ to the true class and equally distributing $\frac{1-\lambda}{k-1}$ to the other classes in the case of k classes. $\tau$ is a label operation that redistributes a portion $\upsilon\in(0,\lambda_2)$ from the true class to the farthest negative class which $\mathcal{N}$ belongs to.  In this manner, linearly scaling perturbation mitigates the high similarity among adversarial samples by overcoming fixed iteration directions in metric-based PGD and diversifying the attack budget. Furthermore, assigning a portion of the labels to farthest negative classes corresponds to the behavior of metric attacks where samples are pulled toward farthest negative classes. This label softening operation facilitates models' learning of robustness-related knowledge about farthest negative classes and alleviates the overfitting problem caused by hard-label based training.


The max-min optimization process for adversarial training based on FNES is formulated as follows:
{
\small
\begin{equation}
\begin{gathered}
\max_{\hat{x}}\mathbb{E}_{\hat{x}}\left [ \mathcal{L}_{\text{metric}}(\hat{x}) \right ],  \\
\min_{G}\mathbb{E}_{(x^{\text{adv}},y^{\text{adv}})}\left [ \mathcal{L}_{\text{cls}}\bigl(G(x^{\text{adv}}),y^{\text{adv}}\bigr)+\mathcal{L}_{\text{tri}}\bigl(E(x^{\text{adv}})\bigr) \right ],  \\
\mathcal{L}_{\text{tri}}(f)=\left[d(f,f^{+})-d(f,f^{-})+m\right]_{+},
\end{gathered}
\label{equation:minmax}
\end{equation}
}
where $\hat{x}$ and $x^{\text{adv}}$ respectively denote the adversarial samples before and after being processed by FNES, $\mathcal{L}_{\text{cls}}$ is the classification loss and $\mathcal{L}_{\text{tri}}$ is the triplet loss with margin value of $m$. $f^+$ and $f^-$ respectively denote the sample features sharing the same ID as $x$ and having a different ID from $x$.

\begin{table*}[h]
\fontsize{7.5}{7}\selectfont  
\centering
\caption{Verification experiment that our method alleviates the challenge of ``partial robustness is distributed on the classifier".}
\setlength{\tabcolsep}{2mm}{
\begin{tabular}{cccccccccc}
\toprule
\multirow{2}{*}{Models}   & \multirow{2}{*}{Datasets} & \multirow{2}{*}{Defense} & \multirow{2}{*}{Clean} & \multicolumn{3}{c}{8/255-16}            & \multicolumn{3}{c}{10/255-16}           \\ \cline{5-10}
                          &                           &                          &                        & FNA         & SMA         & IFGSM       & FNA         & SMA         & IFGSM       \\ \midrule
\multirow{6}{*}{ResNet50} & \multirow{3}{*}{Market}   & AT\_PGD                  & 64.24/85.48            & 26.18/47.86 & 36.32/58.22 & 27.86/50.36 & 19.25/38.39 & 30.72/50.53 & 21.26/41.57 \\
                          &                           & Metric AT                       & 67.20/88.00            & 28.38/52.26 & 45.38/68.74 & 33.97/58.52 & 22.02/42.96 & 40.56/63.18 & 28.57/52.23 \\
                          &                           & \textbf{Ours}                     & \textbf{68.50/88.21}   & \textbf{31.99/55.17} & \textbf{50.13/72.60} & \textbf{37.61/62.02} & \textbf{24.80/45.93} & \textbf{45.96/67.34} & \textbf{31.20/54.22} \\ \cline{2-10}
                          & \multirow{3}{*}{Duke}     & AT\_PGD                  & 53.60/73.70            & 22.34/38.59 & 33.40/50.72 & 26.48/44.03 & 16.69/29.59 & 29.16/45.47 & 21.33/35.61 \\
                          &                           & Metric AT                       & 54.47/74.55            & 23.76/39.27 & 40.40/57.99 & 27.04/44.21 & 17.41/30.97 & 35.98/52.42 & 21.55/36.58 \\
                          &                           & \textbf{Ours}            & \textbf{55.29/75.81}    & \textbf{26.02/43.49} & \textbf{40.74/59.69} & \textbf{31.23/50.67} & \textbf{20.76/36.54} & \textbf{37.50/55.83} & \textbf{26.89/45.29}  \\ \bottomrule
\end{tabular}}
\label{tab:classifier_rob}
\end{table*}
\begin{table*}[h]
\fontsize{7.5}{7}\selectfont  
\centering
\caption{Cross datasets evaluation to verify the generalization ability of our methods on ResNet50.}
\setlength{\tabcolsep}{2.3mm}{
\begin{tabular}{lclccccccc}
\toprule
\multirow{2}{*}{Models}   & \multirow{2}{*}{Datasets}                                                   & \multicolumn{1}{c}{\multirow{2}{*}{Defense}} & \multirow{2}{*}{Clean} & \multicolumn{2}{c}{FNA} & \multicolumn{2}{c}{SMA} & \multicolumn{2}{c}{IFGSM} \\ \cline{5-10}
                          &                                                                             & \multicolumn{1}{c}{}                         &                        & 8/255-16   & 10/255-16  & 8/255-16   & 10/255-16  & 8/255-16    & 10/255-16   \\ \midrule
\multirow{6}{*}{ResNet50} & \multirow{3}{*}{\begin{tabular}[c]{@{}c@{}}Market\\ to\\ Duke\end{tabular}} & None                                         & 15.08/27.65            & 0.15/0.13  & 0.14/0.13   & 0.35/0.36        & 0.28/0.22        & 0.29/0.36         & 0.26/0.36         \\
                          &                                                                             & Metric AT                                           & 16.51/29.35            & 4.47/10.89   & 2.96/9.07        & 10.77/22.52        & 9.49/21.14        & 5.78/13.14         & 4.53/12.07         \\
                          &                                                                             & \textbf{Ours}                                & \textbf{19.07/34.69}   & \textbf{6.17/12.88 }   & \textbf{4.65/10.23}     & \textbf{13.02/24.60}        & \textbf{11.71/22.80}  & \textbf{7.92/15.35}       &  \textbf{6.46/13.33}         \\ \cline{2-10}
                          & \multirow{3}{*}{\begin{tabular}[c]{@{}c@{}}Duke\\ to\\ Market\end{tabular}} & None                                         & 25.18/53.47             & 0.31/0.42        & 0.28/0.27        & 0.76/0.98        & 0.67/0.80        & 0.70/0.98         & 0.62/0.80         \\
                          &                                                                             & Metric AT                                           & 19.54/45.25              & 6.27/17.67        & 4.50/12.80        & 13.32/30.43        & 11.57/26.51        & 7.50/20.61         & 5.89/16.00         \\
                          &                                                                             & \textbf{Ours}                                & \textbf{21.87/50.12}     & \textbf{7.87/21.91}   & \textbf{6.20/18.05}        & \textbf{14.16/34.32}        & \textbf{12.73/30.46}        & \textbf{9.67/25.83}         & \textbf{8.09/21.85}        \\ \bottomrule                          
\end{tabular}}
\label{tab:cross_data_eval}
\end{table*}
\begin{figure*}[h]
\centering
\includegraphics[width=0.65\linewidth,height=4.5cm]{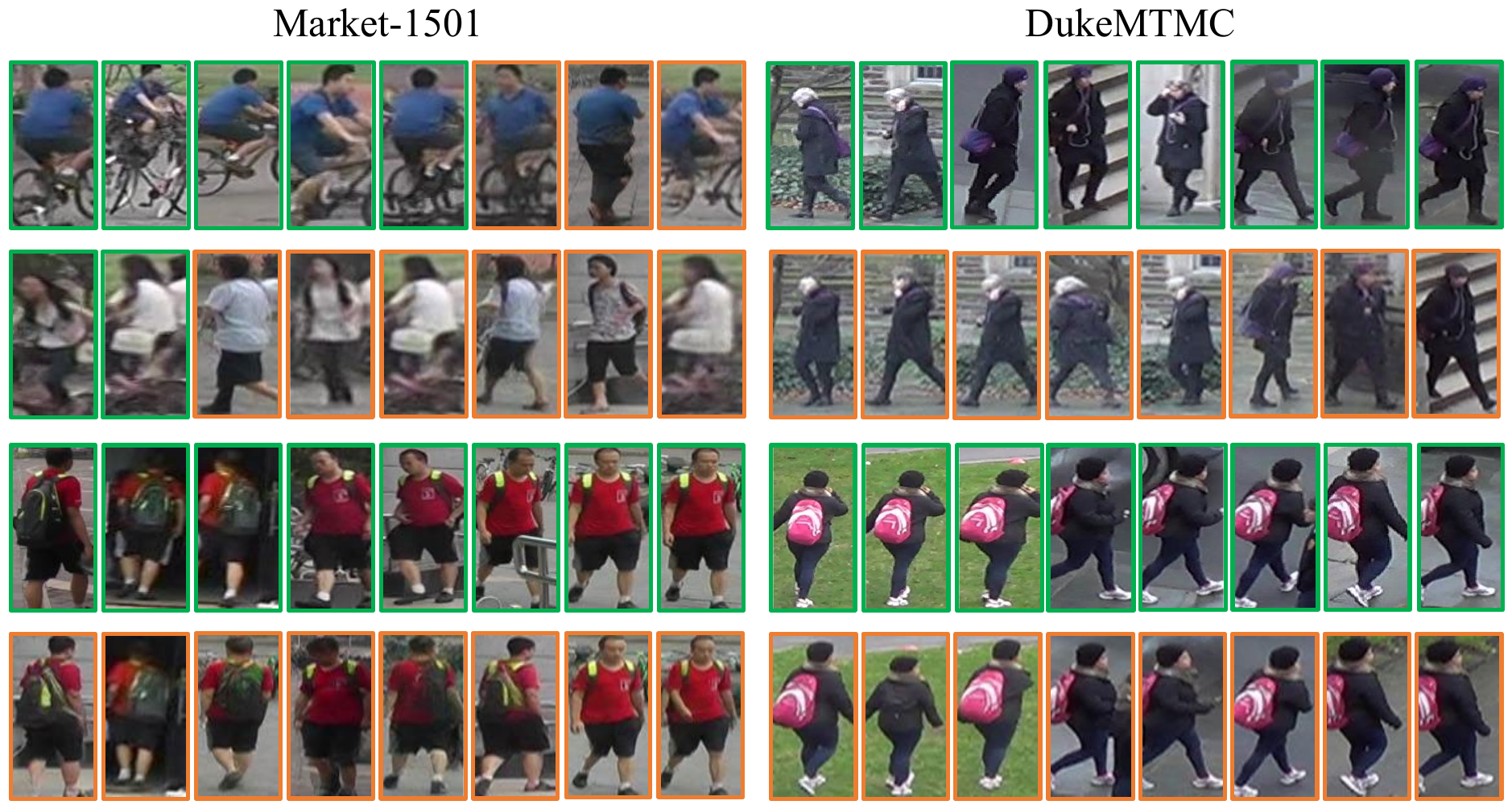}
\caption{Visualization of partial augmented data. The images with green borders are the original samples, while those with orange borders are the generated augmented samples.}
\label{fig:data_aug_vis}
\end{figure*}
\begin{figure*}[h]
\centering
\includegraphics[width = 0.78\linewidth]{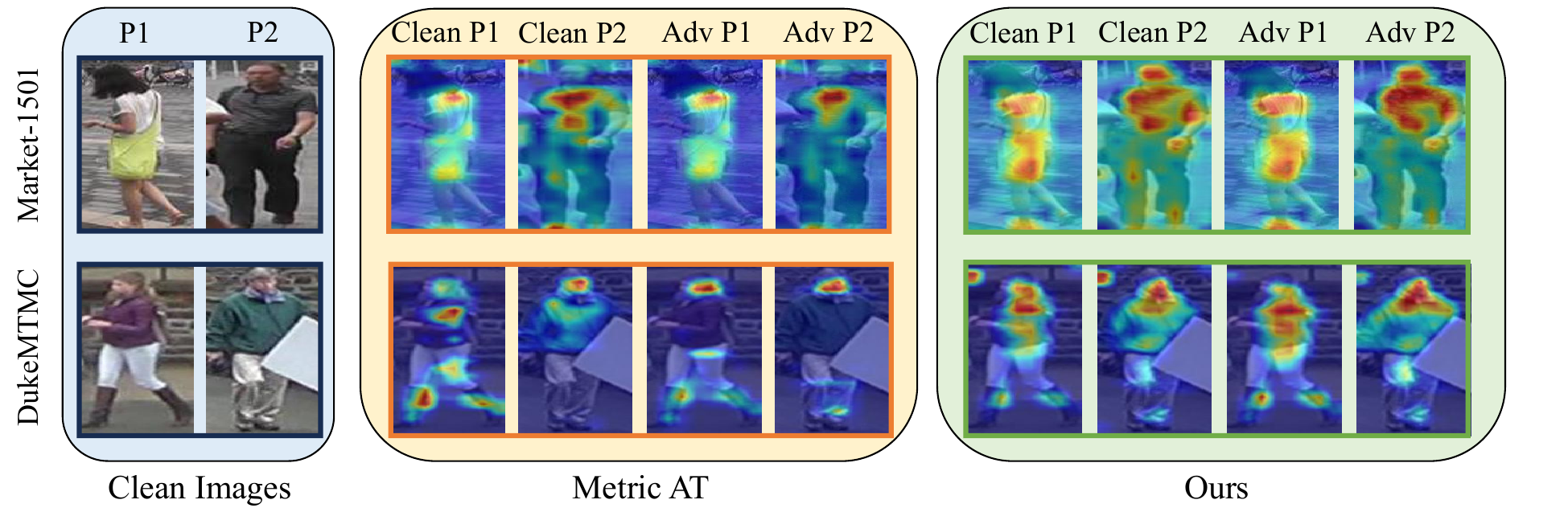}
\caption{Heatmap based Grad-Cam. P1 means the $1_{\text{st}}$ person example.}
\label{fig:gradcam}
\end{figure*}
\begin{figure*}[h]
\centering
\includegraphics[width = 13cm, height=4cm]{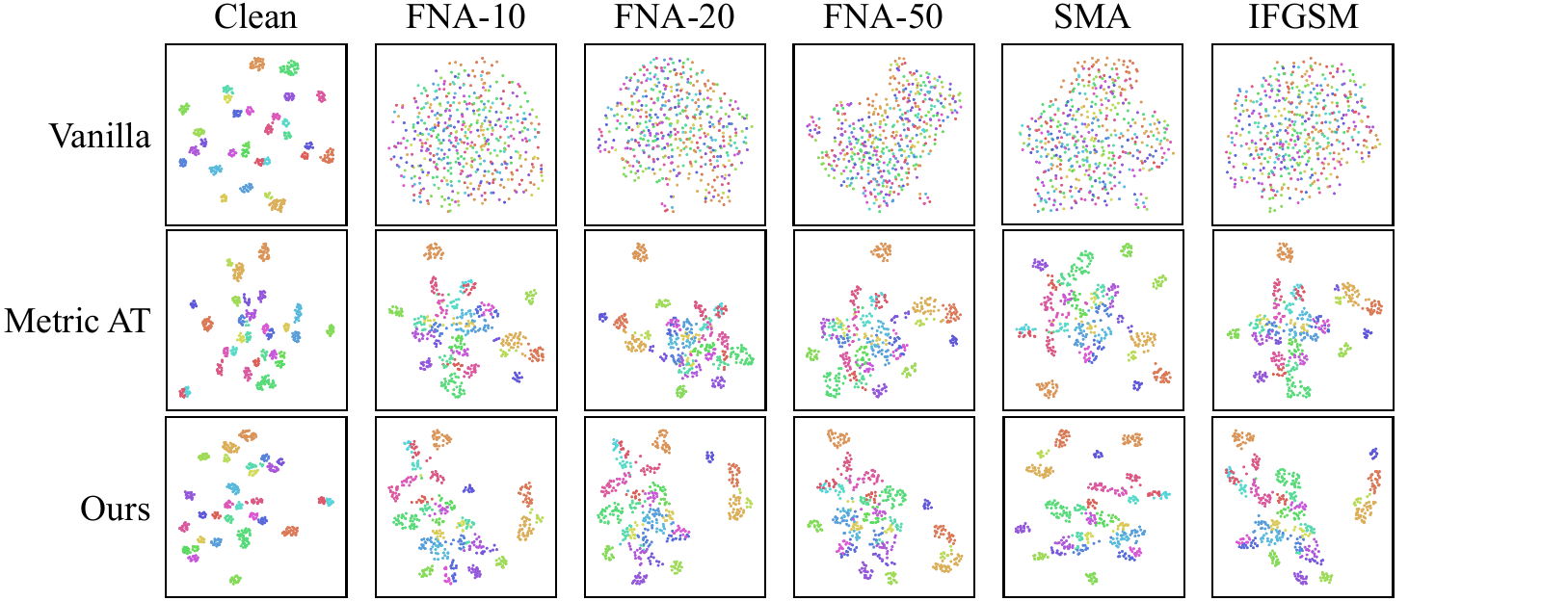}
\caption{Feature distribution visualization based on UMAP.}
\label{fig:umap}
\end{figure*}

\textit{Adversarially-enhanced learning.} To learn adversarial-invariant features that are shared between clean samples and adversarial samples, we introduce a feature discriminator $D$ parameterized by $\theta_{D}$, which, together with the feature encoder $E$, forms an adversarial learning framework. Specifically, the discriminator \( D \) is trained to distinguish whether the features originate from $x^{\text{adv}}$ or $x$, the loss of which can be formalized as:
{\small
\begin{equation}
\mathcal{L}_D=-\mathbb{E}_{x}[\log D(E(x))]-\mathbb{E}_{x^{\text {adv }}}\left[\log \left(1-D\left(E\left(x^{\text {adv }}\right)\right)\right)\right].
\end{equation}}

The objective of $E$ is to maximally confuse the discriminator $D$, thereby preventing $D$ from determining whether the output features originate from $x$ or $x^{\text{adv}}$. This objective can be formalized as maximizing the loss of $D$:
{
\small
\begin{equation}
\mathcal{L}_E=\mathbb{E}_{x}[\log D(E(x))]+\mathbb{E}_{x^{\text{adv}}}\left[\log \left(1-D\left(E\left(x^{\text{adv}}\right)\right)\right)\right].
\end{equation}
}

Ideally, the feature encoder $E$ and the feature discriminator $D$ should finally achieve the Nash equilibrium. At this point, the discriminator perceives all encoded features as either clean or adversarial features with probability of 0.5, meaning it cannot distinguish the origin of features at all. And the encoder extracts adversarial-invariant features shared between adversarial samples and clean samples. The above optimization process can be summarized as a min-max optimization framework:
{
\small
\begin{equation}
\begin{aligned}
\min_E \max_D \mathcal{L}(E,D) & =  \mathbb{E}_x[\log D(E(x))] \\  & +  \mathbb{E}_{x^{\text {adv }}}\left[\log \left(1-D\left(E\left(x^{\text {adv }}\right)\right)\right)\right],
\end{aligned}
\end{equation}
}
where $D$ and $E$ are alternately optimized until $D$ cannot distinguish features (i.e., Nash equilibrium).

\textit{Self-meta learning.}
To learn the generalized invariant features shared between seen and unseen IDs, we introduce self-meta learning. Specifically, for a batch of clean samples $\tilde{x}$ from the target distribution $\mathcal{D}$ and their adversarial samples $\tilde{x}^{\text{adv}}$, we divide them into a simulated training distribution $\mathcal{D}_{\text{meta-train}}$ for meta-training process and a simulated testing distribution $\mathcal{D}_{\text{meta-test}}$ for meta-testing process. The model first computes the loss function on $\mathcal{D}_{\text{meta-train}}$, and then updates its parameters via one-step gradient descent to obtain a temporary model $G_{\text{temp}}$:
{
\small
\begin{equation}
\theta_{G}^{\text{temp}}=\theta_{G} - \alpha \nabla_{\theta_{G}} \mathcal{L}_{\text {meta-train }}\left(\mathcal{D}_{\text {meta-train }}\right),
\end{equation}
}
where $\alpha$ is the temporary learning rate, and $\mathcal{L}_{\text {meta-train }}$ is defined as:
{
\small
\begin{equation}
\mathcal{L}_{\text {meta-train }}=\mathbb{E}_{(x, y) \sim \mathcal{D}_{\text {meta-train }}}[\ell(G, x, y)+ \ell(G, x^{\text{adv}}, y^{\text{adv}})].
\end{equation}
}
where $x^{\text{adv}}$ and $y^{\text{adv}}$ is the corresponding adversarial sample and label of $x$ and $y$. And $\ell=\mathcal{L}_{\text{cls}}+\mathcal{L}_{\text{tri}}+\mathcal{L}_{E}$ is the total loss function. Subsequently, we use the same loss function as the meta-training phase to evaluate $G_{\text{temp}}$ on the meta-testing data and obtain the meta-test loss:
{
\small
\begin{equation}
\mathcal{L}_{\text {meta-test}}=\mathbb{E}_{(x, y) \sim \mathcal{D}_{\text {meta-test }}}\left[\ell\left(G_{\text {temp}},x, y\right)+ \ell(G_{\text {temp }},x^{\text{adv}}, y^{\text{adv}})\right].
\end{equation}
}

Finally, by integrating the losses from the meta-training and meta-testing phases, we obtain the overall loss for the self-meta learning:
{
\small
\begin{equation}
\mathcal{L}_{\text {self-meta learning}}=\mathcal{L}_{\text {meta-train }} + \mathcal{L}_{\text {meta-test}}.
\end{equation}
}

Notice that gradient descent is applied directly to $\theta_G$ based on  $\mathcal{L}_{\text{self-meta learning}}$, instead of sequentially performing gradient descent on the meta-training data followed by the meta-testing data. The optimization can be formalized as:
{
\small
\begin{equation}
\begin{gathered}
\theta_G \leftarrow \theta_G-\beta \nabla_{\theta_G} \mathcal{L}_{\text{self-meta learning}}(\theta_G)  \\
\Updownarrow \\
\min _{\theta_G}\left[\mathcal{L}_{\text{meta-train}}(\theta_G)+\mathcal{L}_{\text{meta-test}}\left(\theta_G-\alpha \nabla_{\theta_G} \mathcal{L}_{\text{meta-train}}(\theta_G)\right)\right],
\end{gathered}
\end{equation}
}
where $\beta$ is the final learning rate. Intuitively, the meta-learning process evaluates whether the update direction of the meta-training phase has adaptive capability for the meta-testing data. If the adaptive capability is poor, the loss function value of the intermediate single-step updated model parameter $\theta^G_{\text{temp}}$ on the meta-testing data will be large, and this larger meta-testing loss corrects the update direction in return for better adaptation ability.

Additionally, the inherent second-order optimization scheme of meta-learning implicitly regularizes the model's first-order gradients, resulting in better generalization. To illustrate, we use the gradient chain rule to expand the gradient of the loss function in the meta-testing phase as an explicit function of the initial parameters $\theta_{G}$:
{
\small
\begin{equation}
\begin{aligned}
& \nabla_{\theta_G} \mathcal{L}_{\text {meta-test }}\left(\theta^{\text {temp }}_{G}\right) 
\\  & =\nabla_{\theta^{\text {temp }}_{G}} \mathcal{L}_{\text {meta-test }}\left(\theta^{\text {temp }}_{G}\right) \cdot \nabla_{\theta_{G}}\left(\theta_{G}-\alpha \nabla_{\theta_{G}} \mathcal{L}_{\text {meta-train }}(\theta_{G})\right)
\\ & =  \underbrace{\nabla_{\theta^{\text {temp }}_{G}} \mathcal{L}_{\text {meta-test }}}_{\text {first-order gradient}}-\alpha \underbrace{\nabla_{\theta_{G}}^2 \mathcal{L}_{\text {meta-train }} \cdot \nabla_{\theta_{G}^{\text {temp }}} \mathcal{L}_{\text {meta-test }}}_{\text {hessian matrix}},
\end{aligned}
\end{equation}
}
where the second-order term $\nabla^2_{\theta_{G}}\mathcal{L}_{\text{meta-train}}$ implicitly regularizes the changing rate of the first-order gradient $\nabla_{\theta_{G}}\mathcal{L}_{\text{meta-train}}$ as:
{
\small
\begin{equation}
\begin{aligned}
&\text{Parameter update direction} \\ & \propto \nabla_{\theta_G} \mathcal{L}_{\text {meta-train }}-\alpha\nabla_{\theta_G}^2 \mathcal{L}_{\text {meta-train }}\cdot \nabla_{\theta^{\text {temp }}_G} \mathcal{L}_{\text {meta-test }}.
\end{aligned}
\end{equation}
}

Here, the second-order term penalizes the curvature of the loss function (i.e., the rapid changes in the gradient), forcing the optimization path to be smoother and avoiding sharp local extrema, thus bringing better generalization.

\subsection{Overall Model}
By integrating the data balancing and the bi-adversarial self-meta defense, we obtain the final model. The intuitive framework can be referenced in Figure~\ref{fig:model}. The pseudocode of our method is provided in the supplementary materials.

\section{Experiments}
\subsection{Basic Setup}
\textit{Datasets and backbones.} Following DAS~\cite{wei2024towards}, two common datasets, DukeMTMC and Market-1501, are used for evaluation. For the backbone networks, to ensure consistency, we also select ResNet50~\cite{he2016deep} and APNet~\cite{chen2021person}. Due to page limitations, please refer to the supplementary material for all evaluations of APNet. Following DAS, our adversarial training is conducted with FNA attack~\cite{bouniot2020vulnerability}.

\textit{Evaluation metrics.} Following the common setup of the person ReID~\cite{zhou2023stochastic,wei2024towards}, we report the mean average precision (mAP) and Rank-1 accuracy of cumulative match characteristic (CMC). FNA, SMA~\cite{bouniot2020vulnerability} and Metric IFGSM~\cite{bai2020adversarial} attacks are used for robustness evaluation. Table values follow the ``mAP/Rank-1'' format.

\textit{Compared defense.}
Due to the fundamental differences between classification tasks and retrieval tasks, most defense methods suitable for classification tasks do not apply to retrieval tasks. Thus, following DAS~\cite{wei2024towards}, we compare our method with the following defenses: BitSqueezing~\cite{xu2018feature}, MedianSmoothing2D~\cite{xu2018feature}, AMD~\cite{bai2020adversarial}, Adv\_train~\cite{bouniot2020vulnerability}, and DAS. 


\textit{Training and evaluation details.} Under our experimental setup, we run each experiment three times and report the best-performing result. For details regarding the software and hardware environments, hyperparameters of each module, attack settings, and training configurations, please refer to the supplementary materials.

\subsection{Main Results}
We report the main white-box robustness comparison on ResNet50 in Table~\ref{tab:white_box_resnet50}. The results used for comparison are referenced from DAS~\cite{wei2024towards}. Our method achieves the current optimal adversarial robustness on the two datasets. To demonstrate its transferability, we also evaluate the black-box robustness in the supplementary material, which proves its transferable robustness.

\subsection{Ablation Analysis}
Firstly, we briefly analyze the effectiveness of each module based on ResNet50. Specifically, when none of our proposed modules is used, the method degenerates to FNA-based metric adversarial training baseline which is denoted as Metric AT. We first analyze the performance of each module when functioning individually, and then, we combine all modules to analyze the final model. The ablation study results are summarized in Table~\ref{tab:abla}. Each module contributes a significant incremental benefit, and the optimal results are obtained when all modules are integrated.

\begin{figure}[t]
  \centering
  \includegraphics[width=0.75\linewidth]{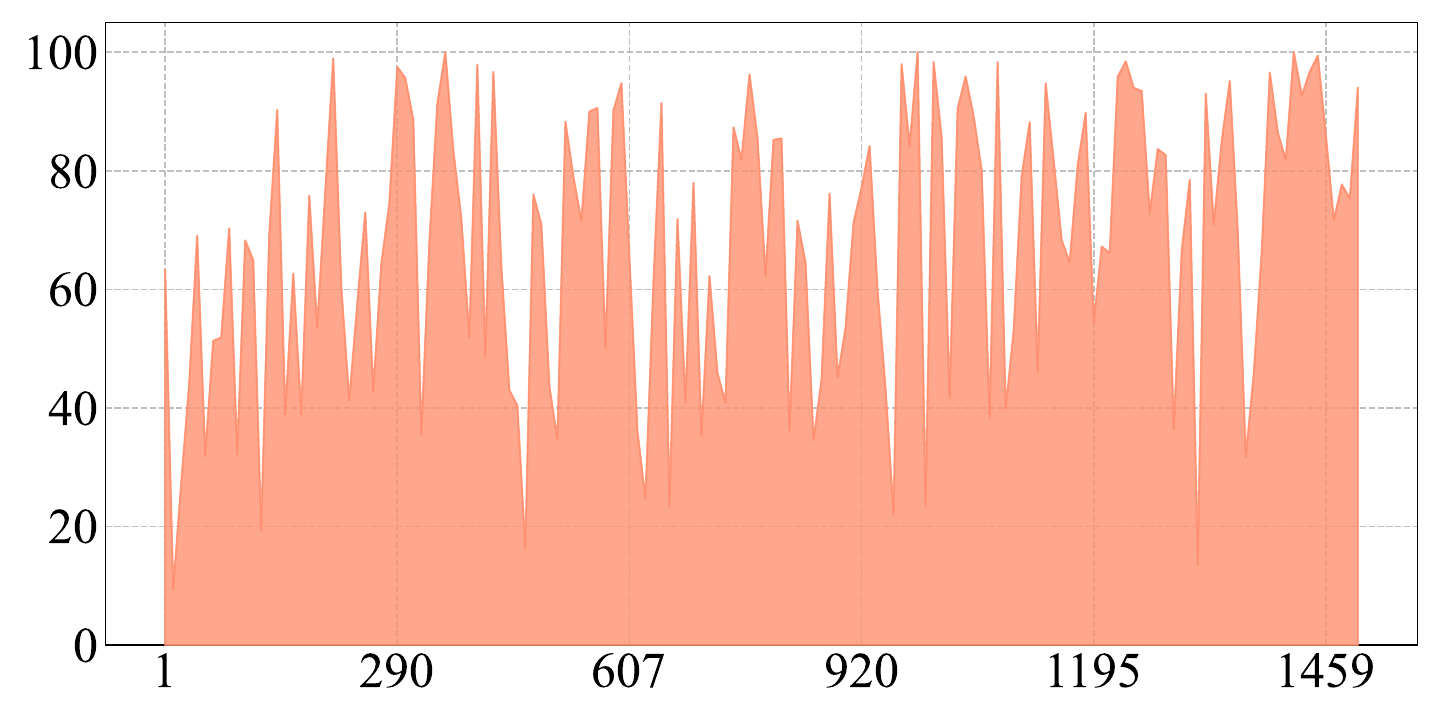}
  \caption{The mAP for each ID (avg=67.59, std.=22.74).}
  \label{fig:each_ID_map_ours}
\end{figure}

Secondly, we follow the similar idea as in the introduction to verify that our method mitigates the challenges of model bias and robustness generalization. For the model bias challenge, the augmented data are shown in Figure~\ref{fig:data_aug_vis}, where IDs with fewer intra-class samples are filled to an average sample number level, and IDs lacking diversity are filled with isomeric pseudo data. We also statistically analyze the accuracy of our method across each IDs on Market-1501, as shown in Figure~\ref{fig:each_ID_map_ours}. Compared to Metric AT, our method exhibits lower inter-ID accuracy variance. For the generalization challenge, our method shows better robustness than Madry's AT~\cite{madry2018towards} denoted as AT\_PGD which uses PGD-based classification attack for adversarial training (shown in Table~\ref{tab:classifier_rob}). This proves that our method alleviates the impact of absent classifiers. The evaluation of different attack methods in Table~\ref{tab:white_box_resnet50} preliminarily validates the generalization ability against unseen attacks. We also reference evaluation protocols for domain adaptation~\cite{mekhazni2020unsupervised,feng2021complementary} and domain generalization~\cite{lin2020multi,liu2024domain} to assess the generalization of our method on unseen domains and IDs. Specifically, we evaluate the robustness of the model trained on dataset A (e.g., Market) when tested on dataset B (e.g., Duke). Results shown in Table~\ref{tab:cross_data_eval} demonstrate significant superiority of our defense on unseen domains, as the goals of adversarially-enhanced learning and self-meta learning modules explicitly induce the model to learn the  dual-invariant representation.

\subsection{Qualitative Analysis}
We further validate the effectiveness of the proposed method through qualitative analyses in this section.

\textit{Heatmap.} The heatmap reflects high attention areas of the model to the samples. A model with strong discriminative ability should focus on key regions within the samples, such as the contours and body of a person. We visualize the heatmaps of both the Metric AT model and our proposed model using Grad-CAM~\cite{selvaraju2017grad}, as shown in Figure~\ref{fig:gradcam}. Visually, our method restores the model's attention areas on adversarial samples, focusing on regions with higher discriminative information. The heatmap of our method is more interpretable than that of the Metric AT model since it is more in line with human contours. This provides interpretability of adversarial robustness from the perspective of the model's preferences.

\textit{Feature distribution visualization.} Intuitively, a model with strong discriminative ability should encode features with smaller intra-class distances and larger inter-class distances. We use UMAP~\cite{mcinnes2018umap} to visualize the feature space distributions of the vanilla trained model, the Metric AT model, and our proposed defense model in Figure~\ref{fig:umap}. Visually, the encoding features of the vanilla model for adversarial samples are scattered in the feature space, indicating poor discriminability. The Metric AT model maintains a discernible feature distribution for adversarial samples and our method demonstrates tighter intra-ID clustering and greater inter-ID distance compared to Metric AT. This provides the interpretability on adversarial robustness from the perspective of feature distribution.

\section{Conclusion}
In this paper, we analyze the two challenges ignored by current adversarial defenses in person ReID, i.e., model bias and composite generalization requirements. We then propose a novel debiased dual-invariant defense to address these challenges. we address model bias through data balancing and address composite generalization requirements through a bi-adversarial self-meta defense framework. Our defense achieves the optimal person ReID adversarial robustness, and the extended experiments and ablation analysis demonstrate its effectiveness, transferability, and interpretability. We hope our work can draw attention to trustworthy and robust person ReID models within the community.

\section{Acknowledgments}
This work was supported in part by the National Natural Science Foundation of China 62572150 and 62372137, in part by the Shenzhen Science and Technology Program KJZD20230923114806014 and JCYJ20230807094411024, in part by the Guangdong Basic and Applied Basic Research Foundation 2024A1515012299, and in part by the Major Key Project of PCL PCL2024A05.

\bibliography{arxiv}

\end{document}


\maketitle

\begin{table*}[h]
\normalsize
\centering
\caption{White-box robustness results on APNet.}
\setlength{\tabcolsep}{0.3mm}{
\begin{tabular}{cclccccccc}
\toprule
\multirow{2}{*}{Model}     & \multirow{2}{*}{Dataset} & \multicolumn{1}{c}{\multirow{2}{*}{Defense}} & \multirow{2}{*}{Clean} & \multicolumn{2}{c}{FNA}  & \multicolumn{2}{c}{SMA}   & \multicolumn{2}{c}{IFGSM} \\ \cline{5-10} 
                           &                          & \multicolumn{1}{c}{}                         &                        & 8/255-16    & 10/255-16  & 8/255-16    & 10/255-16   & 8/255-16    & 10/255-16   \\ \midrule
\multirow{16}{*}{APNet}    & \multirow{8}{*}{Market}  & Origin                                       & 78.52/92.10            & 0.72/0.68   & 0.69/0.71  & 1.00/1.00   & 0.66/0.65   & 2.30/2.93   & 2.14/2.58   \\
                           &                          & BitSqueezing                                 & 77.89/92.10            & 0.95/1.21   & 0.86/0.97  & 1.90/1.90   & 1.46/1.36   & 3.13/3.85   & 3.08/4.09   \\
                           &                          & MedianSmoothning                             & 78.22/91.42            & 2.39/2.96   & 2.14/2.79  & 10.29/13.59 & 10.51/13.09 & 7.21/10.42  & 6.55/9.47   \\
                           &                          & BS+MS                                        & 78.52/92.01            & 3.19/4.39   & 2.74/3.65  & 10.49/12.52 & 9.83/11.22  & 8.28/12.08  & 8.05/11.78  \\
                           &                          & AMD                                          & \textbf{78.77/92.37 }           & 0.75/0.69   & 0.70/0.72  & 1.46/1.43   & 1.06/0.89   & 2.66/3.68   & 2.50/3.30   \\
                           &                          & Adv\_train                                   & 70.41/88.93            & 9.67/17.45  & 5.44/9.82  & 21.31/31.17 & 15.37/22.65 & 18.35/32.83 & 12.77/23.84 \\
                           &                          & DAS                                          & 71.89/89.04            & 14.14/23.72 & 8.15/13.71 & 30.94/44.03 & 24.44/35.06 & 22.15/36.40 & 15.03/25.38 \\
                           &                          & \textbf{Ours}                                         & 73.26/90.23            & \textbf{37.39/60.04} &\textbf{ 29.66/50.83}   & \textbf{59.10/79.63}         & \textbf{55.05/75.33}
         & \textbf{44.40/66.78}         & \textbf{37.88/59.89}         \\ \cline{3-10}
                           & \multirow{8}{*}{Duke}    & Origin                                       & \textbf{67.84/84.25}            & 0.42/0.53   & 0.39/0.53  & 1.22/1.34   & 1.04/0.94   & 2.04/3.05   & 1.72/2.28   \\
                           &                          & BitSqueezing                                 & 67.67/83.80            & 0.52/0.49   & 0.46/0.35  & 1.78/1.92   & 1.59/1.84   & 2.33/3.32   & 2.01/2.42   \\
                           &                          & MedianSmoothning                             & 67.16/83.08            & 1.45/1.79   & 1.23/1.52  & 6.60/9.02   & 5.91/7.67   & 5.45/8.43   & 4.69/7.27   \\
                           &                          & BS+MS                                        & 67.16/83.08            & 1.45/1.79   & 1.23/1.52  & 6.60/9.02   & 5.91/7.67   & 5.45/8.43   & 4.69/7.27   \\
                           &                          & AMD                                          & 66.74/83.08            & 1.62/2.24   & 1.33/1.84  & 7.72/10.09  & 6.67/9.11   & 5.88/9.78   & 5.49/8.30   \\
                           &                          & Adv\_train                                   & 59.15/78.01            & 7.75/14.31  & 4.20/8.03  & 18.39/26.75 & 12.82/19.34 & 14.37/24.68 & 10.18/18.04 \\
                           &                          & DAS                                          & 60.58/78.19            & 13.03/22.21 & 7.31/13.06 & 26.50/39.67 & 20.31/30.92 & 18.76/32.00 & 13.10/22.89 \\
                           &                          & \textbf{Ours}                                         & 60.41/78.99            & \textbf{29.55/47.40} & \textbf{23.09/38.46} & \textbf{46.41/65.35} & \textbf{43.56/62.66} & \textbf{36.98/56.24} & \textbf{31.93/50.76}         \\ \bottomrule
\end{tabular}}
\label{tab:white_box_apnet}
\end{table*}

\begin{table*}[t]
\normalsize
\centering
\caption{Cross datasets evaluation to verify the generalization ability of our method on APNet.}
\setlength{\tabcolsep}{1mm}{
\begin{tabular}{lclccccccc}
\toprule
\multirow{2}{*}{Models}   & \multirow{2}{*}{Datasets}                                                   & \multicolumn{1}{c}{\multirow{2}{*}{Defense}} & \multirow{2}{*}{Clean} & \multicolumn{2}{c}{FNA} & \multicolumn{2}{c}{SMA} & \multicolumn{2}{c}{IFGSM} \\ \cline{5-10}
                          &                                                                             & \multicolumn{1}{c}{}                         &                        & 8/255-16   & 10/255-16  & 8/255-16   & 10/255-16  & 8/255-16    & 10/255-16   \\ \midrule
\multirow{6}{*}{APNet} & \multirow{3}{*}{\begin{tabular}[c]{@{}c@{}}Market\\ to\\ Duke\end{tabular}}    & None                                         & 20.13/34.56           & 2.36/4.49  & 1.95/3.32  & 0.41/0.76        & 0.30/0.31        & 0.59/1.08        & 0.51/1.03        \\
                          &                                                                             & Metric AT                                           & 20.19/36.31           & 6.44/13.06  & 3.73/9.29  & 12.39/22.62        & 10.39/18.40        & 8.52/17.01        & 5.76/12.33          \\
                          &                                                                             & \textbf{Ours}                                & \textbf{21.54/37.02}  & \textbf{7.68/14.48}  & \textbf{4.65/9.74}  & \textbf{13.55/24.24}        & \textbf{12.15/21.63}        & 9.58/17.71        & 6.85/12.66          \\ \cline{2-10}
                          & \multirow{3}{*}{\begin{tabular}[c]{@{}c@{}}Duke\\ to\\ Market\end{tabular}} & None                                         & 27.47/58.49           & 0.62/1.19  & 0.57/1.07   & 1.16/1.93        & 0.91/1.16        & 1.14/2.26        & 1.03/1.96          \\
                          &                                                                             & Metric AT                                           & 21.58/50.56           & 7.42/23.17  & 5.57/18.48  & 14.47/34.43        & 12.84/31.36        & 9.40/28.04        & 7.62/23.64          \\
                          &                                                                             & \textbf{Ours}                                & \textbf{22.58/51.99}  & \textbf{8.77/24.88}  & \textbf{7.00/20.01}  & \textbf{15.05/35.81 }       & \textbf{13.76/32.66}        & \textbf{10.71/28.95}        & \textbf{9.08/24.94}         \\ \bottomrule
                          
\end{tabular}}
\label{tab:cross_data_eval}
\end{table*}

\begin{table*}[t]
\normalsize
\centering
\caption{Black-box robustness. The ``\textit{Source Model}" is the attack source model to generate the adversarial examples.}
\setlength{\tabcolsep}{1mm}{
\begin{tabular}{ccccccccc}
\toprule
\multirow{3}{*}{Dataset} & \multirow{3}{*}{Source Model} & \multirow{3}{*}{Defense} & \multicolumn{3}{c}{ResNet}              & \multicolumn{3}{c}{APNet}               \\ 
                          &                               &                          & \multicolumn{3}{c}{8/255-16}            & \multicolumn{3}{c}{8/255-16}            \\ \cline{4-9}
                          &                               &                          & FNA         & SMA         & IFGSM       & FNA         & SMA         & IFGSM       \\ \midrule
\multirow{12}{*}{Market}  & \multirow{2}{*}{ResNet18}     & Metric AT                       & 65.21/85.87 & 65.61/85.93 & 65.35/85.99 & 71.37/89.64 & 71.64/89.90 & 71.46/89.73 \\
                          &                               & \textbf{Ours}            & \textbf{66.93/87.02} & \textbf{67.32/87.20} & \textbf{67.03/87.20} & \textbf{72.58/89.99} & \textbf{72.91/90.05} & \textbf{72.70/89.96} \\
                          & \multirow{2}{*}{ResNet50}     & Metric AT                       & 64.01/85.15 & 65.09/85.60 & 64.35/85.42 & 70.66/89.37 & 71.25/89.52 & 70.83/89.37 \\
                          &                               & \textbf{Ours}            & \textbf{65.79/86.37} & \textbf{66.88/86.91} & \textbf{66.07/86.40} & \textbf{71.81/89.61} & \textbf{72.46/89.82} & \textbf{72.03/89.67} \\
                          & \multirow{2}{*}{ResNet101}    & Metric AT                       & 64.83/85.57 & 65.36/85.99 & 64.98/85.81 & 71.13/89.58 & 71.52/89.79 & 71.26/89.61 \\
                          &                               & \textbf{Ours}            & \textbf{66.54/86.91} & \textbf{67.13/87.08} & \textbf{66.70/87.08} & \textbf{72.32/89.85} & \textbf{72.71/89.93} & \textbf{72.40/89.85} \\
                          & \multirow{2}{*}{OSNet}        & Metric AT                       & 65.28/85.90 & 65.58/86.07 & 65.31/85.96 & 71.38/89.76 & 71.64/89.85 & 71.40/89.79 \\
                          &                               & \textbf{Ours}            & \textbf{67.02/87.20} & \textbf{67.33/87.32} & \textbf{67.01/87.05} & \textbf{72.61/89.93} & \textbf{72.91/90.11} & \textbf{72.66/90.11} \\
                          & \multirow{2}{*}{VIT}          & Metric AT                       & 60.54/83.05 & 62.28/84.26 & 60.38/82.78 & 67.83/87.71 & 69.16/88.45 & 67.69/87.59 \\
                          &                               & \textbf{Ours}            & \textbf{62.65/84.26} & \textbf{64.27/85.12} & \textbf{62.55/84.44} & \textbf{68.75/88.03} & \textbf{70.16/88.63} & \textbf{68.64/88.06} \\
                          & \multirow{2}{*}{APNet}        & Metric AT                       & 64.83/85.42 & 65.51/86.02 & 64.94/85.69 & 70.75/89.25 & 71.53/89.79 & 70.82/89.25 \\
                          &                               & \textbf{Ours}            & \textbf{66.56/86.76} & \textbf{67.24/87.14} & \textbf{66.67/86.91} & \textbf{71.99/89.64} & \textbf{72.77/90.02} & \textbf{72.06/89.61} \\ \cline{2-9}
\multirow{12}{*}{Duke}    & \multirow{2}{*}{ResNet18}     & Metric AT                       & 53.16/74.24 & 53.29/74.24 & 53.22/74.37 & 56.67/75.80 & 56.81/76.06 & 56.73/75.97 \\
                          &                               & \textbf{Ours}            & \textbf{53.82/74.91} & \textbf{53.99/74.73} & \textbf{53.89/74.82} & \textbf{59.31/77.56} & \textbf{59.41/77.89} & \textbf{59.38/77.74} \\
                          & \multirow{2}{*}{ResNet50}     & Metric AT                       & 53.27/73.34 & 53.90/73.79 & 53.53/73.43 & 57.00/76.08 & 57.48/76.66 & 57.18/76.44 \\
                          &                               & \textbf{Ours}            & \textbf{53.98/73.92} & \textbf{53.60/74.33} & \textbf{54.20/74.19} & \textbf{58.65/77.60} & \textbf{59.08/77.51} & \textbf{58.82/77.65} \\
                          & \multirow{2}{*}{ResNet101}    & Metric AT                       & 52.90/73.97 & 53.17/74.28 & 53.01/74.06 & 56.51/75.84 & 56.75/76.02 & 56.58/75.93 \\
                          &                               & \textbf{Ours}            & \textbf{53.61/74.37} & \textbf{53.87/74.64} & \textbf{53.72/74.55} & \textbf{59.13/77.65} & \textbf{59.31/77.83} & \textbf{59.21/77.67} \\
                          & \multirow{2}{*}{OSNet}        & Metric AT                       & 53.10/74.10 & 53.29/74.19 & 53.14/74.15 & 56.68/75.93 & 56.82/75.97 & 56.73/76.20 \\
                          &                               & \textbf{Ours}            & \textbf{53.86/74.82} & \textbf{54.01/74.73} & \textbf{53.90/74.78} & \textbf{59.25/77.69} & \textbf{59.45/77.74} & \textbf{59.31/77.74} \\
                          & \multirow{2}{*}{VIT}          & Metric AT                       & 50.65/71.77 & 51.61/73.29 & 50.48/71.41 & 54.51/73.69 & 55.22/74.09 & 54.41/73.42 \\
                          &                               & \textbf{Ours}            & \textbf{51.41/72.67} & \textbf{52.25/73.38} & \textbf{51.39/72.58} & \textbf{56.85/75.54} & \textbf{57.72/76.53} & \textbf{56.79/75.63} \\
                          & \multirow{2}{*}{APNet}        & Metric AT                       & 53.42/73.47 & 54.12/74.15 & 53.65/73.65 & 56.80/75.99 & 57.64/76.84 & 57.10/76.53 \\
                          &                               & \textbf{Ours}            & \textbf{54.16/74.06} & \textbf{54.86/74.69} & \textbf{54.32/74.19} & \textbf{58.40/77.11} & \textbf{59.30/77.83} & \textbf{58.66/77.33}  \\ \bottomrule
\end{tabular}}
\label{tab:blackbox}
\end{table*}

\begin{table*}[t]
\normalsize
\centering
\caption{White-box robustness results of CUHK03 and MSMT17 datasets on ResNet50 and APNet.}
\setlength{\tabcolsep}{4mm}{
\begin{tabular}{ccccccc}
\toprule
\multirow{2}{*}{Model}     & \multirow{2}{*}{Dataset} & \multicolumn{1}{c}{\multirow{2}{*}{Defense}} & \multirow{2}{*}{Clean} & \multicolumn{1}{c}{FNA}  & \multicolumn{1}{c}{SMA}   & \multicolumn{1}{c}{IFGSM} \\ 
& \multicolumn{3}{c}{}  & 8/255-16  & 8/255-16  & 8/255-16   \\ \midrule
\multirow{6}{*}{ResNet50}    & \multirow{3}{*}{CUHK03}  & Origin                                            & \textbf{42.98/44.43} & 0.10/0.00 & 0.36/0.07 & 0.23/0.00   \\
                           &                          & Metric AT                & 19.55/16.07 & 4.39/3.07 & 13.15/11.00 & 5.70/4.50\\
                           &                          & \textbf{Ours}            & 23.99/21.57 & \textbf{6.05/4.00} & \textbf{19.55/17.79} & \textbf{8.35/5.64}         \\ \cline{3-7}
                           & \multirow{3}{*}{MSMT17}    & Origin                 & \textbf{48.10/76.63}   & 0.13/0.22   & 0.22/0.22   & 1.14/1.76   \\
                           &                          & Metric AT             & 18.78/42.82   & 8.67/23.95   & 13.92/33.71   & 9.63/26.19   \\
                           &                          & \textbf{Ours}            & 23.38/49.78 & \textbf{10.58/26.65} & \textbf{13.62/33.58} & \textbf{11.98/29.20} \\ \cline{2-7}
\multirow{6}{*}{APNet}    & \multirow{3}{*}{CUHK03}  & Origin                                       & \textbf{35.14/35.50}            & 0.09/0.07  & 1.76/1.64   & 0.82/0.43   \\
                           &                          & Metric AT                             & 25.82/24.14   & 6.58/5.07 & 19.40/17.43 & 9.42/7.57   \\
                           &                          & \textbf{Ours}            & 27.77/26.14            & \textbf{7.35/6.14}
                           & \textbf{23.34/20.86}         & \textbf{10.95/8.71}         \\ \cline{3-7}
                           & \multirow{3}{*}{MSMT17}    & Origin                                       & \textbf{31.92/59.25}   & 0.49/1.41   & 0.69/1.47   & 1.38/3.23   \\
                           &                          & Metric AT             & 19.15/41.90   & 8.23/21.96   & 11.97/27.25   & 9.27/23.92   \\
                           &                          & \textbf{Ours}            & 20.60/45.15 & \textbf{9.45/24.55} & \textbf{13.17/30.28} & \textbf{11.34/28.16}         \\ \bottomrule
\end{tabular}}
\label{tab:cuhk03_msmt17}
\end{table*}

\section{Related Work}
\subsection{Person Re-identification}
Person re-identification (ReID) aims to match specific pedestrian targets across camera networks and serves as one of the core tasks in intelligent surveillance systems. Traditional ReID research primarily relies on single-domain supervised learning and employs deep convolutional networks to extract global or local features~\citep{sun2018beyond}, combined with triplet loss~\citep{hermans2017defense} or classification loss to optimize feature discriminability. However, in practical applications, scarce annotated data and scenario variation may cause significant performance degradation. Domain adaptation ReID focuses on knowledge transfer from source domains (with labeled data) to target domains (without labels). Early approaches adopte unsupervised domain adaptation (UDA), such as cluster-based pseudo-label refinement~\citep{gemutual} or adversarial learning~\citep{deng2018image}, which minimizes inter-domain distribution discrepancies to enhance target-domain performance. Subsequent works introduce intermediate domains~\citep{dai2021dual} or cross-domain attribute disentanglement~\citep{choi2020hi} to further alleviate domain shift issues. Domain generalization in person ReID aims to train universal models without target-domain adaptation, typically by learning domain-invariant features from multi-domain data. For instance, QAConv~\citep{wu2021qaconv} constructs cross-domain image-pair similarity metrics, while MetaBIN~\citep{choi2021meta} employs meta-learning to simulate domain-shift scenarios. Additionally, style transfer~\citep{wang2021instance} and feature disentanglement~\citep{jin2021style} are utilized to enhance model generalization capabilities. Cross-modal person ReID primarily addresses matching between visible-light (RGB) and infrared (IR) modalities and is commonly applied in nighttime surveillance. Representative methods include modality-shared feature learning~\citep{zhang2023local}, modality adversarial training~\citep{wang2020cross}, and shared-modality alignment~\citep{lu2020cross}. Recent studies further incorporate textual descriptions~\citep{chen2022tipcb} or cross-modal consistency learning to reduce inter-modal discrepancies. It is worth noting that research on adversarial robustness for person ReID remains relatively limited. For such a practically relevant task, developing effective robustness training strategies holds significant practical importance for the reliable and secure deployment of person ReID models.

\begin{algorithm*}[tb]
\caption{Debiased Dual-Invariant Defense for Person Re-Identification}
\label{alg:algorithm}
\textbf{Require}: Diffusion model trained on corresponding ReID dataset under an ID-conditional setting. \\
\textbf{Input}: Dataset distribution $\mathcal{D}$, the initial ReID model $G=H(E(\cdot))$, and the initial discriminator $D$. \\
\textbf{Parameter}: The ReID model parameters $\theta_{G} = \theta_{E} \cup \theta_{H}$, and the discriminator parameters $\theta_D$\\
\textbf{Output}: Adversarially robust ReID model $G=H(E(\cdot))$.
\begin{algorithmic}[1] 
\STATE Perform data balancing on $\mathcal{D}$ according to Eq.(1) and Eq.(2) through the diffusion model.
\WHILE{not converge}
\STATE Randomly sample a batch clean samples $(X,Y)$, and calculate corresponding adversarial images $\hat{X}$ by metric attack.
\STATE Get the final samples $(X^{\text{adv}},Y^{\text{adv}})$ by performing FNES according to Eq.(4).
\STATE Calculate discriminator loss $\mathcal{L}_D$ on $(X,Y)$ and $(X^{\text{adv}},Y^{\text{adv}})$ according to Eq.(6).
\STATE Update discriminator parameters $\theta_D$ using gradient descent on $\mathcal{L}_D$.
\STATE Partition $(X,Y)$ into $\mathcal{D}_{\text{meta-train}}$ and $\mathcal{D}_{\text{meta-test}}$, and partition $(X^{\text{adv}},Y^{\text{adv}})$ into $\mathcal{D}_{\text{meta-train}}$ and $\mathcal{D}_{\text{meta-test}}$ at the same ratio.
\STATE Calculate the meta-train loss $\mathcal{L}_{\text{meta-train}}$ of $G$ on $\mathcal{D}_{\text{meta-train}}$ according to Eq.(10).
\STATE Perform a single step gradient descent on $\mathcal{L}_{\text{meta-train}}$ for $G$ and get a temporary model $G_{\text{temp}}$ according to Eq.(9):\\
$\theta_{G}^{\text{temp}}=\theta_{G} - \alpha \nabla_{\theta_{G}} \mathcal{L}_{\text {meta-train }}\left(\mathcal{D}_{\text {meta-train }}\right)$.
\STATE Calculate the meta-test loss $\mathcal{L}_{\text{meta-test}}$ of $G_{\text{temp}}$ on $\mathcal{D}_{\text{meta-test}}$ according to Eq.(11).
\STATE Update model $G$ using gradient descent on $\mathcal{L}_{\text{self meta-learning}}=\mathcal{L}_{\text{meta-train}}+\mathcal{L}_{\text{meta-test}}$ according to Eq.(13):\\
${
\small
\begin{gathered}
\theta_G \leftarrow \theta_G-\beta \nabla_{\theta_G} \mathcal{L}_{\text{self-meta learning}}(\theta_G)
\Longleftrightarrow
\min _{\theta_G}\left[\mathcal{L}_{\text{meta-train}}(\theta_G)+\mathcal{L}_{\text{meta-test}}\left(\theta_G-\alpha \nabla_{\theta_G} \mathcal{L}_{\text{meta-train}}(\theta_G)\right)\right],
\end{gathered}
}$
\ENDWHILE
\end{algorithmic}
\end{algorithm*}

\subsection{Adversarial Attack and Defense}
Adversarial attack aims to fool models into making completely erroneous predictions by injecting imperceptible perturbations into input data. Depending on the amount of information accessible to the attacker, adversarial attacks are typically categorized into white-box and black-box settings. In the white-box attack setting, the attacker has full access to the target model, including its architecture, parameters, and gradients. These attacks primarily include gradient-based approaches~\cite{goodfellow2014explaining, dong2018boosting, madry2018towards}, classifier-based methods~\cite{moosavi2016deepfool}, and optimization-based techniques~\cite{carlini2017towards}. In contrast, black-box attacks assume that the attacker has limited or no access to the target model’s internal information. These attacks can be further categorized into score-based attacks~\cite{chen2017zoo}, decision-based attacks~\cite{brendel2018decision}, and transfer-based attacks~\cite{dong2019evading, lin2019nesterov, xie2019improving, zou2020improving}. Among them, due to their practicality and scalability, transfer-based attacks are widely used to evaluate the adversarial robustness of DNNs in the black-box attack setting.

Adversarial defense aims to empower models with correct discriminative capabilities when facing adversarial samples, i.e., adversarial robustness. Early heuristic defenses are proven to be unreliable defenses based on "gradient masking". Adversarial training~\cite{zhang2019theoretically, zhang2024meta, madry2018towards} is recognized as the most reliable empirical defense, where adversarial examples are generated online during training and incorporated into the training set. Robustness distillation~\cite{goldblum2020adversarially, zi2021revisiting} can be regarded as a variant of adversarial training, where a robust large teacher model distills robust knowledge to a smaller student model.

\subsection{Adversarial Attack and Defense for ReID}
Current adversarial attack and defense methods primarily target classification models, making them inevitably suboptimal when directly transferred to ReID tasks. This is mainly because ReID models focus on learning metric relationships rather than classification capabilities. Research on adversarial attack and defense for person ReID models remains relatively limited. AMA~\cite{bai2020adversarial} is proposed as the first metric attack method. Generated adversarial examples are added into the training set to perform offline adversarial training. A multi-stage network architecture~\cite{wang2020transferable} is employed to extract transferable features. Two efficient attack methods,namely Self-Metric Attack (SMA) and Furthest-Negative Attack (FNA), are later proposed, based on which an online adversarial training framework is introduced~\cite{bouniot2020vulnerability}. SMA separates perturbed images from clean images at the feature level, while FNA combines a hard-negative mining strategy to push adversarial samples closer to the most distant negative samples. A recent defense method learns channel-wise and scale-invariant representations from a data augmentation perspective to improve adversarial robustness~\cite{gong2022person}. Despite these initial advances, current adversarial defense methods for ReID predominantly focus on more challenging inner maximization problems, without systematically considering the unique characteristics of ReID tasks and datasets. Consequently, they achieve suboptimal performance.

\section{Detailed Algorithm}
The detailed training algorithm is outlined in Algorithm~\ref{alg:algorithm} using pseudocode.

\begin{table*}[h]
\normalsize
\centering
\caption{Evaluation of our method's effectiveness under different batch sizes.}
\setlength{\tabcolsep}{3mm}{
\begin{tabular}{ccccccc}
\toprule
\multirow{2}{*}{Model}     & \multirow{2}{*}{Defense} & \multicolumn{1}{c}{\multirow{2}{*}{batch size}} & \multirow{2}{*}{Clean} & \multicolumn{1}{c}{FNA}  & \multicolumn{1}{c}{SMA}   & \multicolumn{1}{c}{IFGSM} \\ 
& \multicolumn{3}{c}{}  & 8/255-16  & 8/255-16  & 8/255-16   \\ \midrule
\multirow{8}{*}{ResNet50}    & \multirow{4}{*}{Metric AT}  & 32                                             & 67.06/87.71 & 30.72/55.88 & 47.51/70.40 & 36.90/62.35   \\
                           &                          & 64                & 67.20/88.00 & 28.38/52.26 & 45.38/68.74 & 33.97/58.52\\
                           &                          & 128            & 60.47/83.67 & 25.49/48.43 & 39.54/62.11 & 29.92/53.95         \\
                           &                          & 256            & 57.51/81.29 & 23.12/44.60 & 36.42/59.32 & 27.22/49.67         \\
                           \cline{2-7}
                           & \multirow{4}{*}{Ours}    & 32                       & 68.38/87.86 & 33.43/57.87 & 52.56/74.41 & 39.97/65.29   \\
                           &                          & 64                & 68.50/88.21 & 31.99/55.17 & 50.13/72.60 & 37.61/62.02\\
                           &                          & 128            & 63.81/85.51 & 27.76/49.38 & 45.68/67.67 & 33.41/56.62         \\
                           &                          & 256            & 59.56/83.31 & 26.59/48.75 & 42.94/66.39 & 32.04/55.61 \\ \bottomrule
\end{tabular}}
\label{tab:different_batchsize}
\end{table*}
\begin{table*}[h]
\normalsize
\centering
\caption{White-box robustness results on ResNet101.}
\setlength{\tabcolsep}{3mm}{
\begin{tabular}{ccccccc}
\toprule
\multirow{2}{*}{Model}     & \multirow{2}{*}{Dataset} & \multicolumn{1}{c}{\multirow{2}{*}{Defense}} & \multirow{2}{*}{Clean} & \multicolumn{1}{c}{FNA}  & \multicolumn{1}{c}{SMA}   & \multicolumn{1}{c}{IFGSM} \\ 
& \multicolumn{3}{c}{}  & 8/255-16  & 8/255-16  & 8/255-16   \\ \midrule
\multirow{6}{*}{ResNet101}    & \multirow{3}{*}{Market}  & Origin                                            & \textbf{86.06/94.57} & 0.23/0.24 & 0.70/0.77 & 0.65/0.80   \\
                           &                          & Metric AT                & 69.64/88.66 & 30.25/53.97 & 48.92/70.64 & 36.00/60.65\\
                           &                          & \textbf{Ours}            & 70.18/89.25 & \textbf{32.11/54.93} & \textbf{50.70/72.24} & \textbf{38.03/61.67}         \\ \cline{2-7}
                           & \multirow{3}{*}{Duke}    & Origin                 & \textbf{77.64/88.06}   & 0.20/0.04   & 1.05/1.17   & 0.78/0.99   \\
                           &                          & Metric AT             & 56.52/76.26   & 24.68/42.12   & 37.66/57.11   & 29.08/47.87   \\
                           &                          & \textbf{Ours}            & 57.39/78.28 & \textbf{26.60/44.52} & \textbf{39.97/58.17} & \textbf{31.29/48.74} \\ \bottomrule
\end{tabular}}
\label{tab:resnet101}
\end{table*}
\begin{table}[h]
\normalsize
\centering
\caption{Evaluation of training cost.}
\setlength{\tabcolsep}{2mm}{
\begin{tabular}{cccc}
\toprule
Model          & Dataset  & Defense  & cost (s/epoch)    \\ \midrule
\multirow{6}{*}{ResNet50} &\multirow{3}{*}{Market-1501}  & Origin  & 20.695 \\
                   &      & Metric AT  & 328.5 \\
                   &      & \textbf{Ours}  & 545.5 \\
                   \cline{3-4}
                          &\multirow{3}{*}{DukeMTMC}  & Origin  & 25.5 \\
                   &      & Metric AT  & 420.5 \\
                   &      & \textbf{Ours}  & 627 \\
                   \midrule
\multirow{6}{*}{APNet} &\multirow{3}{*}{Market-1501}  & Origin  & 26.5 \\
                   &      & Metric AT  & 405.5 \\
                   &      & \textbf{Ours}  & 637.5 \\
                   \cline{3-4}
                          &\multirow{3}{*}{DukeMTMC}  & Origin  & 33.5 \\
                   &      & Metric AT  & 519.5 \\
                   &      & \textbf{Ours}  & 774.5 \\
                   \bottomrule
\end{tabular}}
\label{tab:training_cost}
\end{table}

\section{Experiments Settings}
\subsection{Implementation Details}
In this subsection, we illustrate the implementation details for training and inference stage.

\textit{Training stage.} The batchsize of training dataloader is set to 64, and each batch samples 4 images for 16 IDs. All images are uniformly processed into $384\times128$. Both clean and defense models are trained for 120 epochs using Adam optimizer (the initial learning rate is 0.00035), with the MultiStepLR scheduler (the epochs list of learning rate decay is set to $[20,40,60,80,100]$ and the decay factor is set to 0.1). The weight decay values are set to 0.0005 for ResNet50 and 0.001 for APNet. We train all models using a combination of cross-entropy loss and triplet loss, where triplet loss sets margin to 0.3 and uses hard sample mining.

In our farthest negative extension softening, we create adversarial samples by FNA~\cite{bouniot2020vulnerability} with a perturbation budget of 5/255 and 8 iterations, and the other hyper parameters configuration is: $\gamma=1.5$, $\omega$ is sampled from $\mathcal{U}(0.3,0.8)$, $\lambda_1=0.9$ (consistent with vanilla training and standard adversarial training), $\lambda_2=0.95$ and $\upsilon=0.01$. In our adversarial-invariant learning, we set the scaling factor of loss $\mathcal{L}_E$ to 0.001, and train defense models with cross-entropy loss, triplet loss and $\mathcal{L}_E$. The decay factor of the scheduler of the discriminitor $D$ is set to 0.5, the weight decay of $D$ is set to 0.0005, and the other settings are the same as the defense model. In our self-meta learning, the data batch are partitioned into $\mathcal{D}_{\text{meta\_train}}$ and $\mathcal{D}_{\text{meta\_test}}$ with a ratio of 3:1.

\textit{Evaluation stage.} The robustness of all trained models is evaluated with three attack methods: FNA~\cite{bouniot2020vulnerability}, SMA~\cite{bouniot2020vulnerability} and Metric IFGSM~\cite{bai2020adversarial}. The distance metric for attacks is Euclidean distance, the perturbation budget $\epsilon$ is set to 8/255 and 10/255, and the iteration number of attacks is set to 16. During inference stage, Adversarial samples are generated for the query set to evaluate the robustness of models under attacks.

\subsection{Software and Hardware Environments}
All our experiments are conducted on an NVIDIA A6000 GPU with 48 GB memory. We use pytorch 1.12.1 as the deep learning framework, fastreid as the baseline for person ReID, and advertorch to implement various variants of PGD. Under this experimental setup, we ran each experiment three times and reported the best-performing result.

\section{Extension Experiments}

\subsection{Results on APNet}
In addition to the results of ResNet50~\cite{he2016deep} reported in the main text and following the experimental protocol of DAS~\cite{wei2024towards}, we supplement our evaluations with APNet~\cite{chen2021person}. Specifically, Table~\ref{tab:white_box_apnet} presents the white-box attack robustness of APNet, while Table~\ref{tab:cross_data_eval} demonstrates its cross-dataset generalization performance. The results confirm that our method consistently achieves state-of-the-art performance on the APNet backbone.

\subsection{Black-box Robustness}
Furthermore, to validate the transferability of our approach, Table~\ref{tab:blackbox} compares the black-box attack performance between Metric AT and our method. Here, ``Source Model'' refers to the surrogate model (trained via vanilla training) used to generate transferable black-box attacks. The results demonstrate that our method achieves superior black-box robustness across both ResNet50 and APNet backbones on two benchmark datasets.

\subsection{Results of CUHK03 and MSMT17 Datasets}
In addition to the Market-1501~\cite{zheng2017dukemtmc} and DukeMTMC~\cite{zheng2015market} datasets, we evaluate the performance of the ResNet50 and APNet architectures on both the smaller CUHK03~\cite{Li2014cuhk03} dataset and the larger MSMT17~\cite{Wei2018msmt17} dataset. The results, summarized in Table~\ref{tab:cuhk03_msmt17}, further verifies the applicability of our method across datasets of varying scales.

\subsection{Ablation Analysis of Batch Size}
Since the self-meta learning component of our method necessitates the partitioning of each mini-batch into meta-train and meta-test sets, it is crucial to validate whether the batch size impacts its performance. The results presented in Table~\ref{tab:different_batchsize} demonstrate that our method consistently surpasses Metric AT across a range of batch sizes. This ablation analysis uses Market-1501 dataset for the evaluation.

\subsection{Results on ResNet101}
To demonstrate the efficacy of our method across models with varying parameter counts, we compare it with Metric AT using the ResNet101 architecture on Market-1501 and DukeMTMC datasets. As shown in Table~\ref{tab:resnet101}, our method proves effective not only for the smaller ResNet50 but also for the larger ResNet101, thereby affirming its generalizability across model architectures of different sizes.

\subsection{Evaluation of Training Cost}
The training costs of our method, Metric AT, and standard training are evaluated, as summarized in Table~\ref{tab:training_cost}. The results confirm that the cost of our method is comparable to that of Metric AT, indicating no significant additional computational cost.

\bibliography{arxiv}